\PassOptionsToPackage{svgnames}{xcolor}
\PassOptionsToPackage{comma,authoryear,compress}{natbib}
\documentclass[11pt,letterpaper]{mystyle}

\usepackage{xcolor}
\definecolor{CalGoldHex}{HTML}{F2F8FD}
\providecolor{YaleBlue}{RGB}{0,53,107}
\usepackage[all]{hypcap}
\hypersetup{colorlinks=true,citecolor=YaleBlue}
\usepackage{booktabs}
\usepackage{tabularx}
\usepackage{multirow}
\usepackage{array}
\usepackage{placeins}
\usepackage{graphicx}
\usepackage{microtype}
\usepackage{amsmath}
\usepackage{amssymb}
\usepackage{enumitem}
\usepackage{makecell}
\usepackage{wrapfig}
\usepackage{authblk}
\usepackage{pifont}

\titlespacing*{\section}{0pt}{2.2ex plus 0.4ex minus 0.2ex}{0.8ex}
\titlespacing*{\subsection}{0pt}{1.6ex plus 0.3ex minus 0.2ex}{0.5ex}
\titlespacing*{\paragraph}{0pt}{1.0ex plus 0.2ex minus 0.1ex}{0.6em}
\renewcommand{\arraystretch}{1.04}

\newcolumntype{Y}{>{\raggedright\arraybackslash}X}

\makeatletter
\@ifundefined{iflatexml}{\expandafter\newif\csname iflatexml\endcsname}{}
\iflatexml
  \def\authornotes#1{\gdef\HTMLAuthorNotes{#1}}
\fi
\makeatother

\title{FinRiskAtlas: Decision-Aligned Evaluation of Large Language Models for Financial Risk Review}
\runningtitle{Decision-Aligned Evaluation of Large Language Models for Financial Risk Review}

\iflatexml
\author[1,2]{Suyang Zhong\textsuperscript{\textdagger}}
\author[1]{Jingzhe Zhu\textsuperscript{\textdagger}}
\author[1,3]{Qi Xu\textsuperscript{\textdagger}}
\author[1]{Liyao Sun}
\author[4]{Yin Wang}
\author[1]{Qingqing Sun\textsuperscript{*}}
\author[1]{Shuai Chen\textsuperscript{*}}
\author[1]{Tianyi Zhang\textsuperscript{*}}
\else
\author[1,2,$\dagger$]{Suyang Zhong}
\author[1,$\dagger$]{Jingzhe Zhu}
\author[1,3,$\dagger$]{Qi Xu}
\author[1]{Liyao Sun}
\author[4]{Yin Wang}
\author[1,*]{Qingqing Sun}
\author[1,*]{Shuai Chen}
\author[1,*]{Tianyi Zhang}
\fi
\affil[1]{Ant International}
\affil[2]{Xiamen University}
\affil[3]{Shanghai University}
\affil[4]{Tongji University}
\authornotes{%
 \textsuperscript{\textdagger}\,These authors contributed equally to this work.\par
 \textsuperscript{*}\,Corresponding authors to whom correspondence should be addressed.%
 }

\begin{document}

\begin{abstract}
Deploying large language models for professional financial review requires more than measuring general financial competence: models must perform the specific review operation required by a workflow and determine whether the available evidence is sufficient for a defensible decision. Existing financial benchmarks provide broad coverage of knowledge, reasoning, compliance, and professional tasks, but their evaluation units are often organized around datasets or task formulations rather than the professional decisions that deployed systems support. We introduce \textbf{FinRiskAtlas}, a Chinese-language benchmark that evaluates financial LLMs through two complementary dimensions: operation execution under fixed evidence states and evidence-state control under evolving review conditions. The static benchmark contains 9,742 instances across 53 task families, including 42 Domain Knowledge families and eleven downstream review operations defined by explicit evaluation contracts. \textbf{FinRisk-Ask} extends this framework through offline replay of 680 pre-action states from 104 de-identified professional trajectories, where future evidence is withheld during inference and used only to construct expert-verified evidence targets. Across 33 model configurations, operation-level evaluation produces non-redundant model rankings, with a mean pairwise Spearman correlation of 0.42 across downstream operations, and knowledge-based shortlisting can incur up to 18.01 points of regret on individual operations. FinRisk-Ask further shows that entering the Ask branch more frequently does not necessarily yield better request targeting or stronger end-to-end evidence acquisition. These results demonstrate that broad financial capability scores do not fully capture where models are reliable in professional workflows, motivating evaluation units aligned with the decisions and evidence states that deployed systems must support.
\end{abstract}

\maketitle
\iflatexml
  \begin{center}
    \small
    \ifdefined\HTMLAuthorNotes\HTMLAuthorNotes\fi
  \end{center}
\fi

\section{Introduction}
    \begin{figure*}[t]
        \centering
        \includegraphics[width=\textwidth]{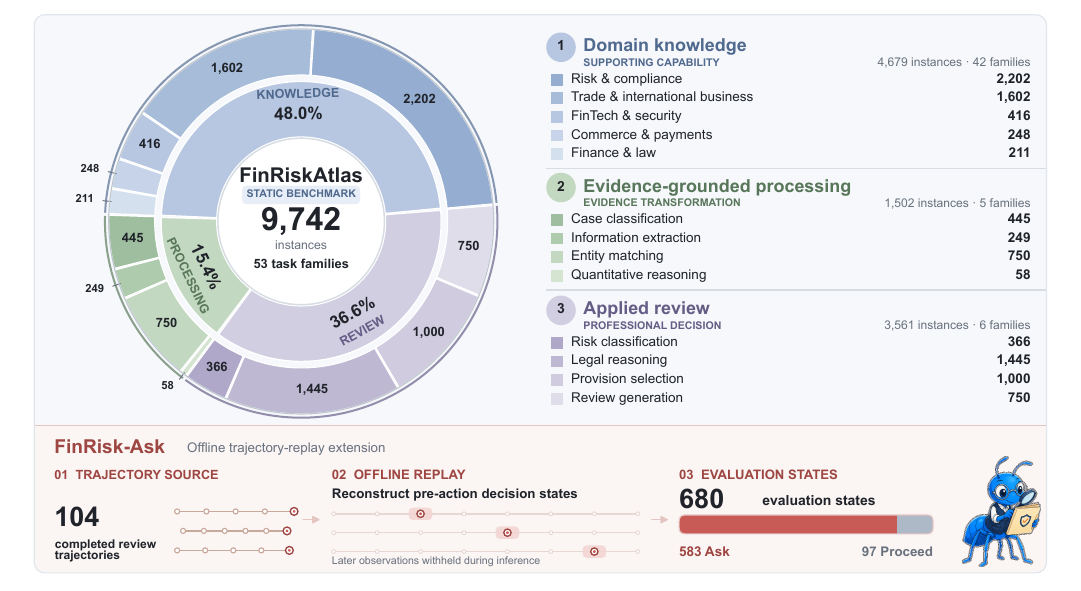}
        \caption{\textbf{Composition of FinRiskAtlas.} The static benchmark evaluates whether a model can execute a specified financial review operation under a fixed evidence state. FinRisk-Ask evaluates whether a model should proceed or acquire additional evidence from an intermediate review state and, after choosing to ask, whether the request targets a trajectory-supported unresolved need. Static instances and trajectory states represent different evaluation units and are evaluated separately.}
        \label{fig:overview}
    \end{figure*}

    Financial risk and compliance review is a sequence of evidence-dependent professional decisions rather than a single prediction task. A reviewer may need to identify transaction parties, extract decision-relevant evidence, verify financial quantities, determine applicable provisions, assign risk categories, or produce a supported professional recommendation. These operations may rely on overlapping records, but they resolve different professional questions and produce different artifacts for downstream decisions. Their reliability also depends on the evidence available at the time of review. For example, when documentation establishing a transaction party's beneficial ownership is missing, a reliable system should recognize that the current evidence state is insufficient and request the necessary information before producing a risk judgment, rather than forcing a conclusion from an incomplete record.
    
    This setting creates two fundamental questions for deploying large language models (LLMs) in professional financial workflows. The first is operation execution: given the currently visible record, which model configuration can reliably perform the required review operation? The second is evidence-state control: given an evolving review state, should the workflow proceed with the available information or acquire additional evidence before making a decision? These capabilities are related but fundamentally different. A model may retrieve the correct regulation yet apply it to the wrong entity, accurately extract requested fields yet fail to integrate them into a case-level assessment, or correctly recognize that information is missing while requesting evidence unrelated to the actual unresolved issue. Therefore, broad financial capability alone does not determine where a model configuration is reliable within a professional review workflow.
    
    Existing financial LLM benchmarks have substantially expanded coverage of financial language understanding and numerical reasoning~\cite{flue,chen-etal-2021-finqa,chen-etal-2022-convfinqa,zhu-etal-2021-tat}, broad financial knowledge, prediction, and professional applications~\cite{xie-etal-2023-pixiu,xie2024holisticfinben,matlin-etal-2025-financial,nie-etal-2025-cfinbench,fineval}, and compliance, safety, and tool-supported research~\cite{ding-etal-2026-cnfinbench,dou-etal-2026-finguard,kim-etal-2026-finred,bigeard-etal-2025-financeagent}. These benchmarks provide valuable measurements of general financial competence. However, their primary evaluation units are commonly organized around knowledge domains, source datasets, or task formulations rather than the professional decisions that models are expected to support in deployment~\cite{xie-etal-2023-pixiu,xie2024holisticfinben,nie-etal-2025-cfinbench,fineval,ding-etal-2026-cnfinbench}. Such units often do not explicitly represent the evidence boundary available to the model, the decision object being resolved, the reviewer artifact being produced, or the criterion by which the output is evaluated. As a result, two tasks derived from the same case record may correspond to different professional operations, while tasks with different surface formats may represent the same operational capability. This creates an evaluation-unit mismatch: a benchmark score may indicate whether a model performs well on a task, but not whether that model is appropriate for a particular decision point in a financial review workflow.
    
    A second limitation concerns the evolution of evidence during professional review. Many fixed-input financial evaluations implicitly assume that the provided record is sufficient for solving the evaluated task~\cite{xie-etal-2023-pixiu,xie2024holisticfinben,nie-etal-2025-cfinbench,fineval}. Real-world financial risk-control processes, however, are iterative: evidence is collected, verified, reconciled, and expanded before decisions are finalized. Prior work on clarification and selective answering~\cite{rao-daume-iii-2018-learning,cole-etal-2023-selectively}, abstention, expert deferral, and help seeking~\cite{feng-etal-2024-dont,mozannar-sontag-2020-defer,ren-etal-2023-ask-help}, and active information acquisition~\cite{li-etal-2024-mediq,li-etal-2025-questbench,zhou-etal-2025-active,zhao-etal-2026-ask} demonstrates the importance of recognizing underspecification. However, recognizing that information is missing is only the first step of evidence acquisition. In professional review, a useful request must identify the specific evidence gap that blocks the current decision, respect the temporal boundary of the review state, and request information that can advance the workflow. Thus, deciding whether to acquire evidence and determining what evidence to acquire represent distinct capabilities.
    
    To address the first limitation, we introduce \textbf{FinRiskAtlas}, a benchmark that organizes financial LLM evaluation around fine-grained professional operations in financial risk-control and compliance review. FinRiskAtlas therefore takes a professional review operation as the downstream evaluation unit, rather than a source dataset or task format. Each downstream family is defined before model evaluation through an explicit contract specifying the visible evidence regime, professional decision object, required reviewer artifact, and scoring protocol. This design allows downstream scores to be interpreted as operation-specific performance signals rather than generic capability measurements. Unlike task-oriented evaluation, operation-oriented evaluation asks whether a model is suitable for the specific decision it is deployed to support. The static FinRiskAtlas benchmark contains 9,742 instances across 53 task families. Forty-two Domain Knowledge families, comprising 4,679 instances, evaluate the concepts, regulations, obligations, and risk mechanisms required for financial risk-control review. Eleven downstream operation families contain 5,063 instances, covering Evidence-Grounded Processing and Applied Review operations, including information extraction, entity matching, quantitative reasoning, risk classification, rule grounding, legal-outcome prediction, and professional review generation. FinRiskAtlas combines broad summary signals with operation-specific evaluation: the Domain Knowledge macro supports coarse screening, while the eleven downstream operations retain their task-native scores for fine-grained configuration selection.
    
    To address the second limitation, we introduce \textbf{FinRisk-Ask}, a complementary benchmark for evidence-state control through offline replay of professional review trajectories. FinRisk-Ask contains 680 pre-action states reconstructed from 104 completed, de-identified trajectories, including 583 recorded Ask states and 97 recorded Proceed states. Candidate models observe only the case record and review history available before the recorded action; later evidence, subsequent reviewer actions, and downstream outcomes are withheld during inference. For each retained Ask state, later-observed evidence is used only to construct expert-verified unresolved evidence targets. Unlike incomplete-information benchmarks constructed from clinical dialogues or deliberately underspecified reasoning problems~\cite{li-etal-2024-mediq,li-etal-2025-questbench,zhou-etal-2025-active,zhao-etal-2026-ask}, FinRisk-Ask grounds request targets in later-observed evidence from completed professional trajectories and evaluates whether a model can acquire the evidence required to advance a professional decision under a realistic information boundary. The evaluation target is not the optimal question in an abstract dialogue, but the evidence request required to advance a concrete professional review decision.
    
    Experiments across 33 model configurations demonstrate why both evaluation dimensions are necessary. The eleven fixed-evidence operations produce non-redundant configuration rankings, with a mean pairwise Spearman correlation of 0.42. Broad knowledge performance does not uniformly preserve operation-specific selection: restricting selection to the Domain Knowledge leader can forgo up to 18.01 points on an individual downstream operation. Evidence-state evaluation reveals another source of divergence. FinRisk-Ask uses ERA as the primary end-to-end measure of the evidence-acquisition pathway: a model receives credit only when it enters the recorded Ask branch and generates a request aligned with an expert-verified unresolved need. ERA is further decomposed into recorded-Ask recall and Conditional Request Alignment, allowing branch-entry failures to be distinguished from request-targeting failures. The results show that similar action-selection behavior can still yield substantially different end-to-end evidence-acquisition performance.
    
    Our contributions are as follows:
    \begin{itemize}[leftmargin=*,nosep]
    \item \textbf{An operation-centric evaluation paradigm for financial review.}
    We introduce FinRiskAtlas, which defines professional review operations as the fundamental evaluation unit for financial LLM assessment. By explicitly modeling the evidence boundary, decision object, reviewer artifact, and scoring protocol of each operation, FinRiskAtlas provides fine-grained evaluation signals aligned with real financial risk-control decisions.
    \item \textbf{A trajectory-grounded benchmark for evidence-state control.}
    We introduce FinRisk-Ask, an offline replay framework that evaluates whether models should proceed or acquire additional evidence and whether their requests target expert-verified unresolved needs, while preventing access to future trajectory information. ERA provides the primary end-to-end measure of the evidence-acquisition pathway, while action- and request-level metrics diagnose where end-to-end performance is gained or lost.
    \item \textbf{Evidence that benchmark granularity affects model selection.}
    Across 33 configurations, we demonstrate that operation-level evaluation and evidence-state evaluation reveal differences hidden by broad capability scores, exposing distinct strengths and failure modes of LLMs in professional financial workflows.
    \end{itemize}
    
\section{Related Work}\label{sec:related-work}

    \subsection{Financial LLM Evaluation: From Capability Coverage to Workflow Operations}
    Financial LLM evaluation has progressed from focused capability assessment to broader measurements of financial knowledge, reasoning, and professional applications. Early benchmarks isolate specific abilities. FLUE evaluates financial language understanding; FinQA and ConvFinQA study numerical reasoning over financial reports and conversational contexts; and TAT-QA combines textual and tabular evidence for financial reasoning~\cite{flue,chen-etal-2021-finqa,chen-etal-2022-convfinqa,zhu-etal-2021-tat}. FinanceReasoning and BizBench further extend evaluation toward complex financial and business reasoning~\cite{tang-etal-2025-financereasoning,krumdick-etal-2024-bizbench}. These benchmarks establish the importance of measuring financial capabilities beyond general-language evaluation.
    
    Subsequent benchmark suites broaden coverage across heterogeneous financial tasks. PIXIU and FinBen evaluate multiple financial capabilities including understanding, reasoning, prediction, and application, while FLaME, CFinBench, and FinEval further expand evaluation of financial knowledge and reasoning, including Chinese financial scenarios~\cite{xie-etal-2023-pixiu,xie2024holisticfinben,matlin-etal-2025-financial,nie-etal-2025-cfinbench,fineval}. Such benchmarks provide valuable capability profiles, but their evaluation units are typically inherited from knowledge categories, source datasets, or task formulations. Consequently, they may not directly correspond to the professional decisions supported by a deployed system. A set of tasks derived from the same case record may represent different review operations, while tasks with different surface formats may serve the same workflow purpose.
    
    Recent work moves closer to realistic financial applications. FinED-Bench evaluates financial document error detection, while Finance Agent Benchmark studies multi-step financial research supported by external tools~\cite{he-etal-2026-large,bigeard-etal-2025-financeagent}. CNFinBench jointly considers financial capability, compliance, and safety; FinGuard derives compliance evaluation from financial regulations; and FinRED studies finance-specific red-team behavior~\cite{ding-etal-2026-cnfinbench,dou-etal-2026-finguard,kim-etal-2026-finred}. These efforts introduce realistic documents, safety considerations, and specialized professional scenarios. However, they generally retain capability- or task-oriented evaluation units, leaving open how benchmark organization should reflect the operational decisions supported by deployed systems.
    
    \subsection{Diagnostic Evaluation and Decision-Aligned Measurement}
    Beyond benchmark coverage, prior work has emphasized the importance of structured diagnostic evaluation. CheckList organizes behavioral tests around specific linguistic capabilities and expected model behaviors, while HELM evaluates models across diverse scenarios and multiple metrics under transparent protocols~\cite{ribeiro-etal-2020-beyond,liang-etal-2023-holistic}. These studies show that aggregate scores alone can hide meaningful differences across capabilities.
    
    Professional-domain benchmarks similarly adopt structured taxonomies to expose specialized abilities. C-Eval organizes evaluation by knowledge subjects, while LawBench, LegalBench, and DiagnosisArena construct domain-specific evaluations for legal and clinical reasoning~\cite{huang2023ceval,fei-etal-2024-lawbench,guha-etal-2023-legalbench,zhu-etal-2026-diagnosisarena}. These benchmarks demonstrate the value of expert-defined capability structures.
    
    However, existing diagnostic frameworks primarily organize evaluation around capabilities or scenarios rather than the professional operations and evidence boundaries involved in deployment. Extending diagnostic evaluation toward workflow-aligned measurement remains an open direction for professional applications~\cite{ribeiro-etal-2020-beyond,liang-etal-2023-holistic,fei-etal-2024-lawbench,guha-etal-2023-legalbench}.
    
    \subsection{From Abstention to Active Evidence Acquisition}
    A separate research line studies model behavior when available information is insufficient for reliable answering. Clarification-question formulation and selective answering methods investigate how models handle ambiguity or unanswerable inputs~\cite{rao-daume-iii-2018-learning,zhang-etal-2024-clamber,zhao-etal-2024-couldve,cole-etal-2023-selectively}. Abstention, expert deferral, and help-seeking approaches similarly study whether models can recognize uncertainty and appropriately transfer or delay decisions~\cite{feng-etal-2024-dont,mozannar-sontag-2020-defer,ren-etal-2023-ask-help}.
    
    Recent benchmarks evaluate information acquisition more directly. MediQ studies follow-up question generation in clinical diagnosis, QuestBench evaluates whether models identify missing variables required for reasoning, and active-reasoning benchmarks examine when and what information should be requested under incomplete conditions~\cite{li-etal-2024-mediq,li-etal-2025-questbench,zhou-etal-2025-active,zhao-etal-2026-ask}. RealFin further studies whether financial questions contain unavailable necessary premises~\cite{dai-etal-2026-realfin}. These works demonstrate that answering a fully specified problem and identifying missing information are distinct abilities.
    
    Trajectory-based approaches provide another perspective by using expert interaction histories. Learn-to-Ask uses sequential expert demonstrations and later observations to learn proactive information-seeking policies~\cite{wei-etal-2026-grounded}. FinRisk-Ask shares the use of completed trajectories and later evidence observations, but differs in objective and evaluation setting. It does not learn a policy or optimize future interaction. Instead, it performs offline evaluation of a recorded professional boundary, measuring whether a model should proceed or acquire evidence and whether the requested evidence addresses an unresolved review need. These directions leave open how evidence acquisition behavior should be evaluated within a professional workflow, where the objective is not only to recognize missing information but also to acquire evidence required for a specific decision. This highlights the need for evaluation frameworks that jointly consider the operation being performed and the evidence state from which that operation is reached.
    
\section{The FinRiskAtlas Benchmark}\label{sec:benchmark}

    \begin{figure*}[t]
        \centering
        \includegraphics[width=0.95\textwidth]{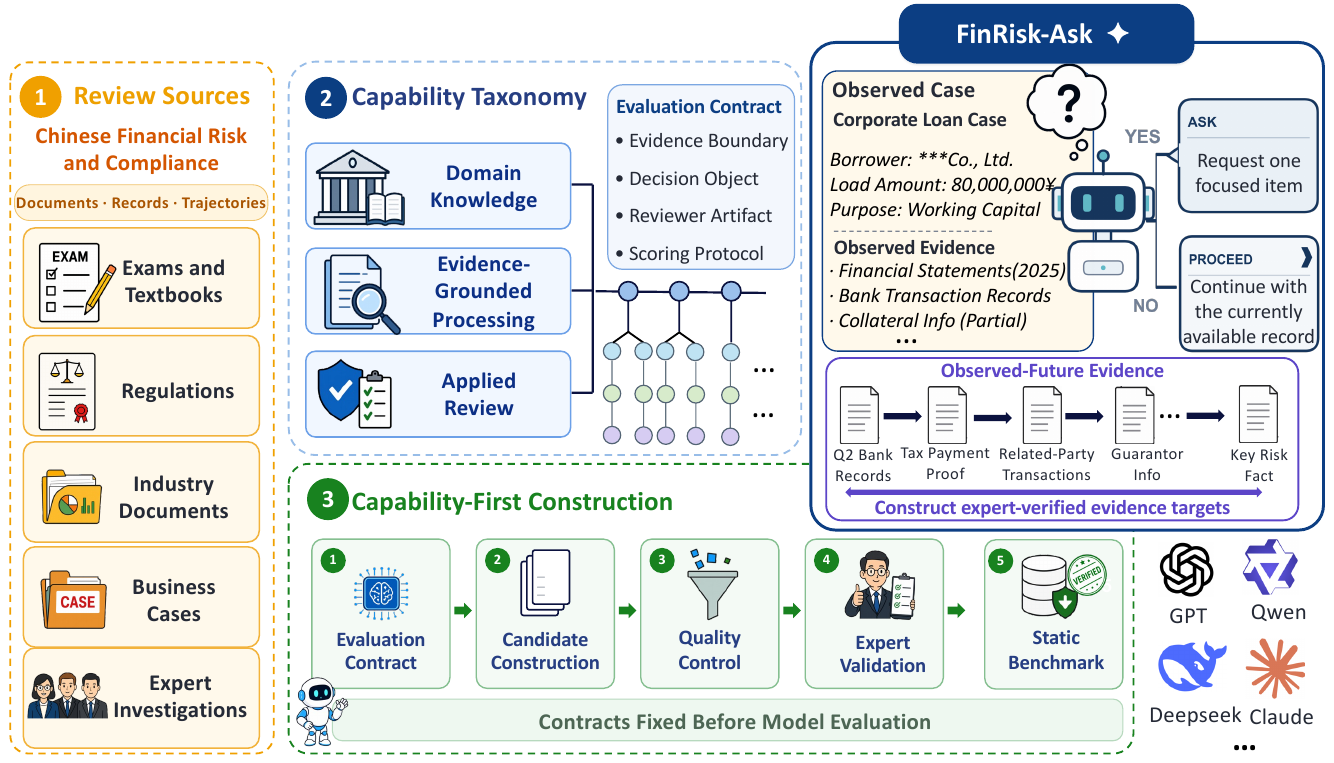}
        \caption{\textbf{Benchmark design around operation and evidence state.} Professional source materials are organized through an operation-aligned taxonomy and fixed evaluation contracts to construct the static benchmark. FinRisk-Ask complements fixed-evidence operation evaluation through offline replay of model-visible pre-action states, measuring both the Ask-or-Proceed transition and the alignment of generated requests with later-observed, expert-verified evidence targets. Later evidence is withheld during inference and used only for target construction.}
        \label{fig:framework}
    \end{figure*}
    
    \subsection{Benchmark Scope and Evaluation Units}

    FinRiskAtlas is built around the observation that model selection in professional workflows occurs at the level of decisions rather than isolated task formats. A useful benchmark unit should therefore correspond to a stable decision context: what information is available to the model, what decision must be resolved, what artifact is expected from the reviewer, and how the result should be evaluated. This design principle motivates operation-centric evaluation, where the benchmark unit reflects the professional action that a model is expected to support. The distinction is important because a deployment decision is not determined by whether a model performs well on an isolated task, but by whether it can reliably produce the artifact required at a specific workflow stage. Changing the evaluation unit changes the selection signal available to practitioners.
    
    Figure~\ref{fig:framework} illustrates the two evaluation settings in FinRiskAtlas. The static benchmark evaluates fixed-evidence review operations, while FinRisk-Ask evaluates control of evolving evidence states. These two settings share the same principle: evaluation should be aligned with the decision context rather than only the surface form of the task.
    
    The static benchmark contains three nested units. A capability layer groups families according to their role in financial review. A task family is the finest stable analysis unit, representing either a review-relevant knowledge area or a concrete fixed-evidence review operation. An instance is one evaluation item governed by the contract of its family. The static benchmark contains 53 families: 42 Domain Knowledge families and eleven downstream operation families. FinRisk-Ask is a separate state-level evaluation setting and is not counted as an additional static family.
    
    To operationalize this principle, we define each family through an explicit evaluation contract:
    \begin{equation}
    \Gamma_f=(c_f,\mathcal{I}_f,d_f,\mathcal{Y}_f,s_f),
    \label{eq:evaluation-contract}
    \end{equation}
    where $c_f$ specifies the capability being measured, $\mathcal{I}_f$ defines the model-visible information regime, $d_f$ specifies the professional decision object, $\mathcal{Y}_f$ defines the required reviewer-artifact schema, and $s_f$ specifies the fixed parsing and scoring protocol. The information regime includes the evidence sources, fields, and temporal boundary available to the model rather than merely the surface format of the input.
    
    The contract defines the measurement boundary rather than a fixed execution order for every review case. Two tasks may originate from the same source case but belong to different families when they resolve different decisions or produce different artifacts. Conversely, records with different surface formats may belong to the same family when they instantiate the same decision context. Therefore, family boundaries are determined by professional function rather than incidental properties of source data.
    
    \subsection{Operation-Aligned Capability Taxonomy}
    
    The static benchmark organizes financial risk-review capability into three layers:
    \[
    \text{Domain Knowledge}
    \rightarrow
    \text{Evidence-Grounded Processing}
    \rightarrow
    \text{Applied Review}.
    \]
    The arrows indicate information dependency rather than a mandatory linear workflow. Individual cases may skip, repeat, or enter conditional branches, and the taxonomy organizes evaluation coverage rather than prescribing the execution order of every professional review process.
    
    Families are grouped according to their role in review. Table~\ref{tab:workflow-map} provides a compact overview of the eleven fixed-evidence downstream operations and the complementary state-level decision evaluated by FinRisk-Ask. Detailed operation contracts, including the model-visible information regime, professional decision object, reviewer artifact, and scoring protocol, are reported in Appendix~\ref{app:protocol}, Table~\ref{tab:operation-contracts}. Domain Knowledge is treated as a supporting layer because its families evaluate concepts, rules, obligations, and risk mechanisms rather than workflow actions. The taxonomy and operation definitions were fixed by domain experts before candidate-model evaluation.
    
    \begin{table*}[t]
        \centering
        \footnotesize
        \setlength{\tabcolsep}{5pt}
        \renewcommand{\arraystretch}{1.12}
        \caption{\textbf{Operation coverage in FinRiskAtlas.}
        The static benchmark contains eleven fixed-evidence review operations.
        FinRisk-Ask evaluates the complementary state-level decision of whether
        the current evidence state should be advanced or expanded.}
        \label{tab:workflow-map}
        \begin{tabularx}{\textwidth}{@{}p{0.21\textwidth}YY@{}}
        \toprule
        \textbf{Evaluation component}
        & \textbf{Operations}
        & \textbf{Workflow outputs}\\
        \midrule
        Evidence-Grounded Processing
        & Case classification; information extraction; institution matching;
          person matching; quantitative reasoning
        & Procedure routing, structured evidence fields, entity-identity
          decisions, and normalized quantities\\
        \addlinespace[3pt]
        Applied Review
        & Risk classification; applicable-provision selection;
          legal-outcome prediction; legal-judgment generation;
          disputed-issue generation; decision-view generation
        & Risk grades, governing provisions, dispositions, reasoned analyses,
          issue lists, and review opinions\\
        \midrule
        \makecell[l]{\textit{State-level extension}\\FinRisk-Ask}
        & Ask-or-Proceed decision, with one focused evidence request after
          selecting Ask
        & Advancement of the current state or a targeted evidence request\\
        \bottomrule
        \end{tabularx}
    \end{table*}
    
    \paragraph{Domain Knowledge.}
    The Domain Knowledge layer evaluates conceptual, regulatory, and risk-related knowledge required for financial risk and compliance review. It contains 42 task families with 4,679 instances distributed across sixteen taxonomy groups, including financial risk, fraud, money laundering, compliance, international trade and finance, financial technology and security, commerce and payments, and finance-related law. These families provide the knowledge foundation required for interpreting evidence and making professional decisions.
    
    \paragraph{Evidence-Grounded Processing.}
    This layer evaluates whether models can transform heterogeneous financial and legal records into structured, decision-relevant evidence. It contains five task families with 1,502 instances: case classification, information extraction, institution matching, person matching, and quantitative reasoning.
    
    \paragraph{Applied Review.}
    This layer evaluates whether models can integrate available evidence, apply relevant rules, and produce professional review outputs. It contains six task families with 3,561 instances: risk classification, legal-outcome prediction, legal-judgment generation, applicable-provision selection, decision-view generation, and disputed-issue generation.
    
    The static benchmark therefore contains 53 families and 9,742 instances. Domain Knowledge is summarized by an unweighted macro-average over its 42 families, while downstream operations retain their task-native metrics. Because downstream operations correspond to different reviewer artifacts and decision objectives, FinRiskAtlas does not collapse them into a single aggregate score.
    
    \subsection{Capability-First Construction and Quality Control}
    
    FinRiskAtlas follows a capability-first construction process. We first define the review-relevant capability or professional operation, establish its evaluation contract, identify suitable source materials, construct candidate instances, and perform expert review. Therefore, family boundaries are determined by intended professional function rather than by available data sources or output formats.
    
    The benchmark draws from four source categories: professional examinations and textbooks; regulatory documents and industry guidance; business, transaction, entity, and legal records; and completed financial-risk review trajectories. Static instances are constructed through three routes: normalization of expert-authored questions, structured transformation of financial and legal records into evidence-processing tasks, and source-conditioned construction of case-level review tasks when the expected output is supported by visible evidence or documented professional decisions.
    
    Importantly, family definitions and evaluation contracts are fixed before candidate-model evaluation. Construction decisions are therefore independent of model predictions and cannot be adapted according to observed model behavior. Six domain experts, including four experts in financial risk and compliance review and two experts in legal review of financial disputes, defined the taxonomy, reviewed family boundaries, audited candidate items, and resolved disagreements through consensus.
    
    Quality control is conducted at both family and instance levels. Family-level review verifies that items within a family share the same capability target, information regime, decision object, reviewer artifact, and scoring protocol. Instance-level review checks evidence support, ambiguity, duplication, information leakage, parser validity, and provenance. Items without a stable reference or reproducible scoring procedure are removed.
    
    \subsection{FinRisk-Ask: Controlling an Evolving Evidence State}
    
    FinRisk-Ask extends the operation-centric evaluation principle to settings where the evidence state itself changes during review. It is not an interactive dialogue benchmark or a policy-learning framework of the kinds studied in prior information-acquisition work~\cite{li-etal-2024-mediq,zhou-etal-2025-active,wei-etal-2026-grounded}; instead, it performs offline evaluation of whether a model can make the appropriate evidence-state transition under a fixed historical decision boundary. The goal is not to optimize future interaction, but to measure whether a model can recognize when additional evidence is required and request evidence that supports a concrete professional decision.
    
    The static benchmark assumes that the visible evidence state is fixed. FinRisk-Ask evaluates whether that state should be advanced or expanded through offline replay of completed, de-identified financial-risk review trajectories collected from an enterprise risk-control workflow.
    
    For a decision point $t$, let $C_t$ denote the case record and review history visible immediately before the recorded action. Candidate models observe only $C_t$; later evidence, subsequent actions, and downstream outcomes are withheld during inference. The model selects either to proceed with the available record or to request additional evidence.
    
    A retained state must satisfy three conditions: the pre-action context can be reconstructed without later information, the immediately following reviewer action maps unambiguously to Ask or Proceed, and no post-action information is included in the model-visible context. Ask states require an additional condition: later evidence must be verified by experts as unavailable at the decision point and relevant to an unresolved review need. Proceed states do not require request targets because their evaluated action is the decision to advance without initiating evidence acquisition.
    
    The released FinRisk-Ask benchmark contains 680 states from 104 trajectories: 583 recorded Ask states and 97 recorded Proceed states. Every released Ask state supports both action evaluation and request evaluation. The reference action represents the recorded professional transition in one workflow; it does not claim to be the unique optimal review policy.
    
    For each retained Ask state, the benchmark constructs an expert-verified set of later-observed evidence needs that were unavailable and unresolved at the replayed decision point. Candidate models never observe these targets during inference. Requests are evaluated against the underlying evidence need rather than the historical wording of the reviewer request, allowing semantically equivalent requests to receive credit.
    
    FinRisk-Ask evaluates end-to-end evidence acquisition on recorded Ask states with ERA, which assigns credit only when a model both enters the Ask branch and requests evidence aligned with a verified unresolved need. CRA complements ERA by isolating request-targeting capability conditional on entering the Ask branch, while BAcc, AskR, and ProceedR characterize agreement with the recorded action boundary and its two branches. Together, these metrics complement static operation-level assessment with a structured evaluation of evidence-state control.

\section{Experiments}\label{sec:experiments}

    \subsection{Experimental Setup}
    
    \paragraph{Evaluation protocol.}
    We evaluate the static FinRiskAtlas benchmark and FinRisk-Ask over 33 model configurations. All candidate evaluations use zero-shot direct-answer inference without in-context demonstrations or requested chain-of-thought. Each model--instance pair is evaluated once under the archived inference configuration, and malformed or unparseable outputs remain in the evaluation denominator. Open-ended Applied Review tasks and FinRisk-Ask request alignment are evaluated with a fixed semantic evaluator configuration. Reported results should therefore be interpreted as point estimates for the evaluated configurations rather than repeated-sampling estimates.
    
    \paragraph{Models and comparison sets.}
    The evaluation suite contains 33 configurations spanning open-weight checkpoints and API systems across multiple scales and model families, including Qwen, DeepSeek, Kimi, GLM, GPT, Gemini, Claude, Nemotron, Ling, Seed, Dianjin, and FinR1. The evaluation unit is the archived configuration rather than an abstract model family because serving configurations and inference settings can affect observed behavior. For analyses involving semantic evaluation, DeepSeek-V4-Flash is excluded from comparative rankings because it also serves as the fixed evaluator. All reported comparative analyses involving judge-scored operations therefore use the remaining 32 configurations. Let $\mathcal{M}_{\mathrm{cmp}}$ denote the 32 non-evaluator configurations used in analyses involving judge-scored tasks. For each downstream operation $o$, let $\mathcal{M}_o$ denote its eligible configuration set: all 33 configurations for the eight structured operations and $\mathcal{M}_{\mathrm{cmp}}$ for the three judge-scored operations. Configurations are ranked within $\mathcal{M}_o$ in descending order of their Domain Knowledge macro score $K_m$. We let $r^K_{m,o}$ denote the resulting competition rank, with rank 1 assigned to the highest score and tied configurations receiving the same rank.

    \begin{figure*}[t]
        \centering
        \includegraphics[width=0.95\textwidth]{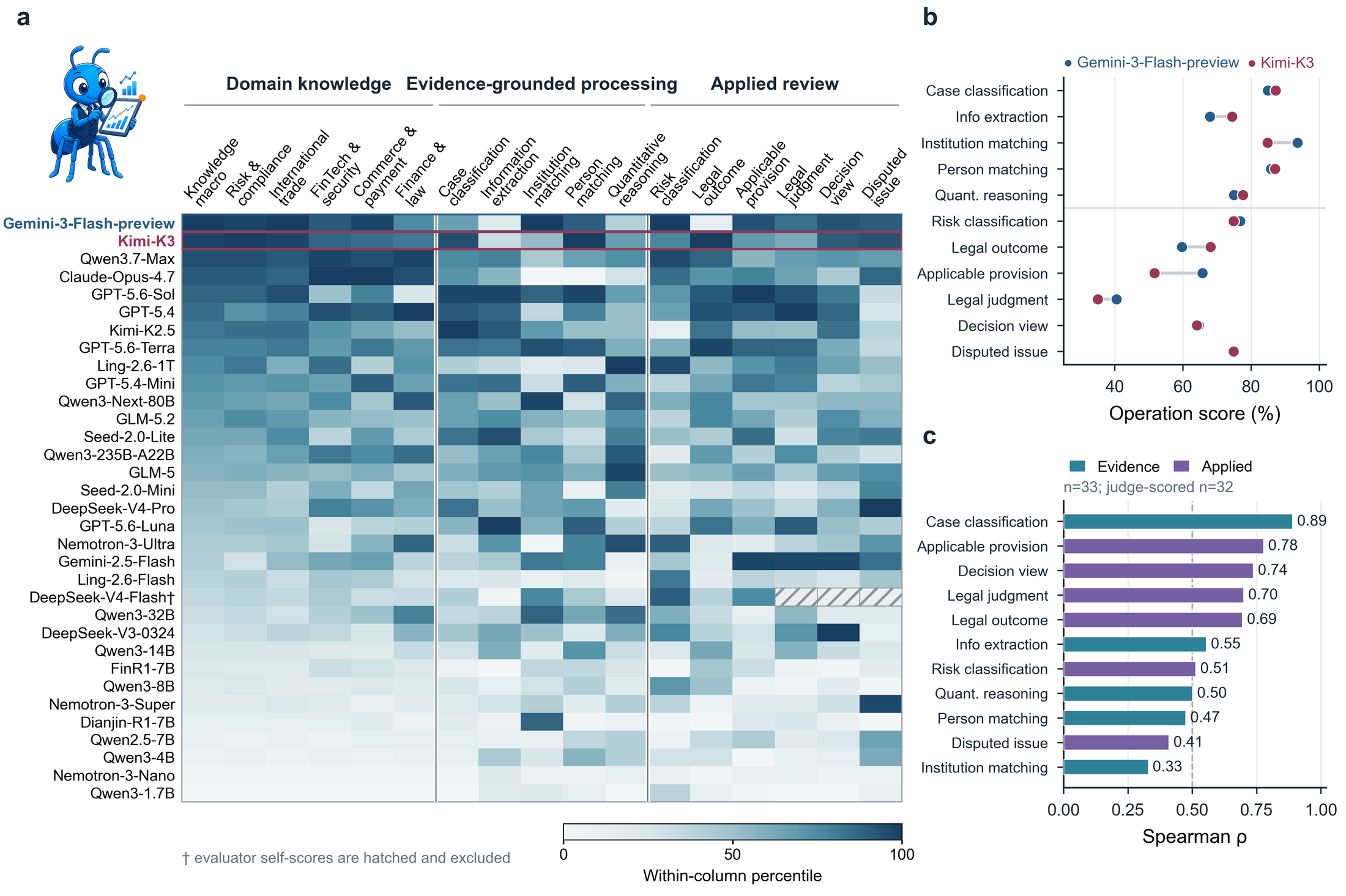}
        \caption{\textbf{Operation-level profiles across evaluated configurations.} (a) Within-column percentile heatmap, with configurations ordered by the Domain Knowledge macro; hatched cells denote evaluator self-scores excluded from evaluator-dependent comparisons. (b) Raw downstream-operation scores for the knowledge-near pair Gemini-3-Flash-preview and Kimi-K3. (c) Spearman correlations between the Domain Knowledge macro and downstream operations, using 33 configurations for structured operations and 32 for judge-scored operations.}
        \label{fig:capability-atlas}
    \end{figure*}
        
    \paragraph{Metrics.}
    For static operations, we retain each operation's native metric because downstream families correspond to different reviewer artifacts and decision objectives. For FinRisk-Ask, Evidence-Request Alignment (ERA) measures end-to-end acquisition on recorded Ask states by jointly accounting for Ask-branch entry and alignment of the resulting request with an expert-verified evidence need. Conditional Request Alignment (CRA) complements ERA by measuring request-targeting capability conditional on entering the Ask branch, making it useful for comparing how effectively different configurations formulate evidence requests. Balanced Recorded-Action Agreement (BAcc), Ask recall (AskR), and Proceed recall (ProceedR) characterize agreement with the recorded action boundary and its two branches. Formal definitions are provided in Appendix~\ref{app:finriskask-protocol}.
    
    \paragraph{Research questions.}
    The experiments investigate three consequences of decision-aligned evaluation. RQ1: How much operation-specific variation exists across downstream review operations? RQ2: How well does broad financial knowledge preserve downstream configuration selection? RQ3: How do action selection and request targeting jointly determine evidence-state performance?
    
    \subsection{RQ1: Operation-Level Evaluation Reveals Heterogeneous Model Strengths}
    
    The downstream review operations induce substantially different configuration rankings despite being evaluated on the same configuration pool. Figure~\ref{fig:capability-atlas}(a) visualizes this behavior by ordering configurations according to their Domain Knowledge macro scores and displaying their within-operation percentile ranks. Rather than preserving a common ordering across downstream operations, many configurations change relative positions between evidence-processing and applied-review tasks, suggesting that different operations emphasize different capabilities.
    
    We quantify this variation by computing the $11\times11$ Spearman rank-correlation matrix over $\mathcal{M}_{\mathrm{cmp}}$ using the complete downstream operation scores (Tables~\ref{tab:full-operation-results} and~\ref{tab:full-generation-results}). The mean pairwise correlation is $\rho=0.42$, with 37 of the 55 operation pairs exhibiting correlations below 0.5. Some operations exhibit almost independent rankings. For example, institution matching and risk classification have a Spearman correlation of only $\rho=-0.03$. The eigenspectrum of the same correlation matrix, shown in Figure~\ref{fig:supplementary-capability-analysis}(b), provides the same conclusion from a complementary perspective. The leading component explains 49.9\% of the total eigenvalue mass, whereas the first five components explain 85.0\%, indicating that downstream operations share a common capability component but cannot be reduced to a single latent ordering.
    
    Figure~\ref{fig:capability-atlas}(c) shows that the correlation between the Domain Knowledge macro and downstream operations varies substantially, ranging from 0.33 for institution matching to 0.89 for case classification. This effect is also visible among configurations with nearly identical knowledge performance. Gemini-3-Flash-preview and Kimi-K3 differ by only 0.01 points on the Domain Knowledge macro, yet Gemini-3-Flash-preview leads by 14.10 points on applicable-provision selection, whereas Kimi-K3 leads by 8.47 points on legal-outcome prediction. Consistent with this observation, knowledge-near configuration pairs (within 0.5 Domain Knowledge points) still exhibit a median downstream profile difference of 25.5 percentile points (Figure~\ref{fig:supplementary-capability-analysis}(a)). Moreover, the eleven downstream operation leaders are distributed across nine different configurations (Figure~\ref{fig:supplementary-capability-analysis}(d)), indicating that no single configuration consistently dominates all review operations. These observations show that operation-level evaluation provides configuration-selection signals that are not preserved uniformly by broad financial capability scores.

    \begin{figure*}[t]
        \centering
        \includegraphics[width=0.95\textwidth]{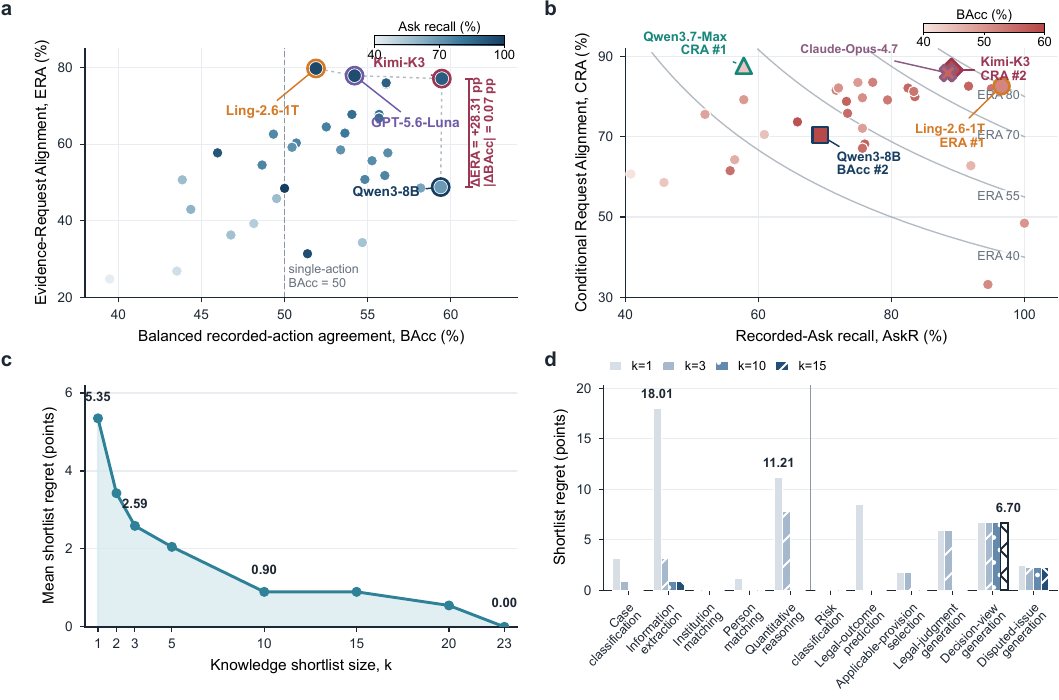}
        \caption{\textbf{Evidence-state behavior and operation-specific screening regret.}
        (a) Relationship between Balanced Recorded-Action Agreement and Evidence-Request Alignment across configurations.
        (b) Decomposition of request alignment into recorded-Ask recall and Conditional Request Alignment.
        (c--d) Score forgone under Domain-Knowledge-based shortlisting across different shortlist sizes.}
        \label{fig:ask-and-regret}
    \end{figure*}
    
    \subsection{RQ2: Knowledge-Based Screening Incurs Operation-Dependent Regret}
    \label{sec:regret}
    
    The heterogeneous rankings observed in RQ1 suggest that knowledge-based screening may not preserve the configurations preferred by individual downstream operations. We therefore quantify the resulting deployment cost by measuring the regret incurred when candidate configurations are shortlisted solely according to their Domain Knowledge ranking.
    
    For operation $o$, let $S_{m,o}$ denote the score of configuration $m$. Restricting selection to the top-$k$ configurations under the Domain Knowledge ranking, the observed shortlist regret is defined as
    
    \begin{equation}
    \mathrm{Reg}_o(k)
    =
    \max_{m\in\mathcal{M}_o}S_{m,o}
    -
    \max_{\substack{m\in\mathcal{M}_o\\r^K_{m,o}\leq k}}
    S_{m,o}.
    \end{equation}

    Ties at the rank threshold are retained, so the set $\{m\in\mathcal{M}_o:r^K_{m,o}\leq k\}$ may contain more than $k$ configurations. This convention avoids arbitrary tie-breaking. Because downstream operations follow heterogeneous evaluation contracts, regret is reported in each operation's native score units. Figure~\ref{fig:ask-and-regret}(c,d) summarizes the resulting operation-specific regret curves. The cost of knowledge-based screening varies substantially across downstream operations. At $k=1$, selecting only the Domain Knowledge leader incurs no regret for institution matching and risk classification, but forfeits 18.01 points on information extraction, 11.21 points on quantitative reasoning, and 8.47 points on legal-outcome prediction. Increasing the shortlist reduces, but does not eliminate, these gaps. For example, decision-view generation still retains 6.70 points of regret even at $k=15$. These results suggest that the Domain Knowledge macro is well suited for coarse screening, whereas selecting the final deployment configuration benefits from operation-level evaluation.
    
    \subsection{RQ3: Evidence-State Control Separates Action Selection from Request Targeting}
    \label{sec:active-review}
    Evidence-state control requires both selecting the appropriate evidence-state transition and generating a request that targets the unresolved evidence need. FinRisk-Ask evaluates these behaviors through complementary metrics. ERA measures the end-to-end acquisition outcome on recorded Ask states, BAcc measures agreement with the full recorded Ask-or-Proceed boundary, and AskR and CRA expose branch entry and conditional request alignment.
    
    Recorded-action agreement does not determine end-to-end evidence acquisition. Figure~\ref{fig:ask-and-regret}(a) shows that, among configuration pairs whose BAcc values differ by no more than 0.1 points, ERA differs by as much as 28.31 points. Configurations can therefore reproduce the recorded action boundary at similar rates while differing substantially in whether their behavior culminates in a request aligned with the unresolved evidence need. Figure~\ref{fig:ask-and-regret}(b) makes the decomposition of ERA explicit by plotting recorded-Ask recall against Conditional Request Alignment and overlaying iso-ERA contours. Configurations in the upper-right region combine broad coverage of recorded Ask states with strong request targeting and therefore achieve high end-to-end alignment. Ling-2.6-1T attains the highest ERA by combining an AskR of 96.57\% with a CRA of 82.58\%. Kimi-K3 and Claude-Opus-4.7 occupy a similar high-coverage and high-alignment region, yielding ERA scores of 77.10\% and 75.98\%, respectively. By contrast, Qwen3.7-Max achieves the highest CRA at 87.67\% but enters the recorded Ask branch on only 57.80\% of recorded Ask states, limiting its ERA to 50.68\%. The panel therefore shows that strong conditional request alignment alone is insufficient for end-to-end evidence acquisition: high ERA requires both reliable entry into the Ask branch and a request that targets the unresolved evidence need.
    
    The same separation is visible in the aggregate action behavior. Across all 32 non-evaluator configurations, AskR exceeds ProceedR. The median Ask--Proceed recall gap is 46.26 points, and 24 configurations exhibit gaps larger than 20 points. This systematic asymmetry indicates that the evaluated configurations reproduce recorded requests for additional evidence more readily than recorded decisions to proceed with the current record. It also motivates reporting BAcc together with the two branch-specific recalls rather than relying on overall action agreement alone. Taken together, ERA provides the primary end-to-end measure of evidence acquisition on recorded Ask states, while BAcc characterizes agreement with the full recorded action boundary and AskR, ProceedR, and CRA diagnose where performance is gained or lost. These results demonstrate that evidence-state control is not a single capability, but the combination of selecting an appropriate evidence-state transition and acquiring evidence that resolves the underlying professional decision.

\section{Conclusion}
FinRiskAtlas studies financial LLM evaluation through the lens of professional decision making. It evaluates whether a model configuration can perform the required review operation and control the evidence state from which that decision is made. The static benchmark provides operation-aligned evaluation under fixed evidence conditions, while FinRisk-Ask extends evaluation to evolving states where models must determine whether additional evidence is needed and whether the requested evidence addresses the underlying decision need. Our experiments show that these two dimensions reveal capability differences hidden by broad benchmark scores. Different review operations induce distinct model profiles, and evidence-state evaluation separates recognizing the need for information from acquiring the right information. These findings suggest that selecting LLMs for professional financial workflows requires evaluation units aligned with deployment decisions rather than relying only on general financial competence. FinRiskAtlas provides the contracts, review operations, reconstructed evidence states, and reproducible evaluation protocols needed to study this perspective. We hope this perspective encourages future benchmarks to align evaluation units more closely with the professional decisions that deployed language models are expected to support.

\newpage
\bibliographystyle{plainnat}
\bibliography{main}

\begin{thebibliography}{38}
\providecommand{\natexlab}[1]{#1}
\providecommand{\url}[1]{\texttt{#1}}
\expandafter\ifx\csname urlstyle\endcsname\relax
  \providecommand{\doi}[1]{doi: #1}\else
  \providecommand{\doi}{doi: \begingroup \urlstyle{rm}\Url}\fi

\bibitem[Bigeard et~al.(2025)Bigeard, Nashold, Krishnan, and Wu]{bigeard-etal-2025-financeagent}
Antoine Bigeard, Langston Nashold, Rayan Krishnan, and Shirley Wu.
\newblock Finance agent benchmark: Benchmarking llms on real-world financial research tasks.
\newblock arXiv preprint arXiv:2508.00828, 2025.
\newblock URL \url{https://arxiv.org/abs/2508.00828}.

\bibitem[Chen et~al.(2021)Chen, Chen, Smiley, Shah, Borova, Langdon, Moussa, Beane, Huang, Routledge, and Wang]{chen-etal-2021-finqa}
Zhiyu Chen, Wenhu Chen, Charese Smiley, Sameena Shah, Iana Borova, Dylan Langdon, Reema Moussa, Matt Beane, Ting-Hao Huang, Bryan Routledge, and William~Yang Wang.
\newblock {F}in{QA}: A dataset of numerical reasoning over financial data.
\newblock In \emph{Proceedings of the 2021 Conference on Empirical Methods in Natural Language Processing}, pages 3697--3711. Association for Computational Linguistics, 2021.
\newblock \doi{10.18653/v1/2021.emnlp-main.300}.
\newblock URL \url{https://aclanthology.org/2021.emnlp-main.300/}.

\bibitem[Chen et~al.(2022)Chen, Li, Smiley, Ma, Shah, and Wang]{chen-etal-2022-convfinqa}
Zhiyu Chen, Shiyang Li, Charese Smiley, Zhiqiang Ma, Sameena Shah, and William~Yang Wang.
\newblock {C}onv{F}in{QA}: Exploring the chain of numerical reasoning in conversational finance question answering.
\newblock In \emph{Proceedings of the 2022 Conference on Empirical Methods in Natural Language Processing}, pages 6279--6292. Association for Computational Linguistics, 2022.
\newblock \doi{10.18653/v1/2022.emnlp-main.421}.
\newblock URL \url{https://aclanthology.org/2022.emnlp-main.421/}.

\bibitem[Cole et~al.(2023)Cole, Zhang, Gillick, Eisenschlos, Dhingra, and Eisenstein]{cole-etal-2023-selectively}
Jeremy Cole, Michael Zhang, Daniel Gillick, Julian Eisenschlos, Bhuwan Dhingra, and Jacob Eisenstein.
\newblock Selectively answering ambiguous questions.
\newblock In \emph{Proceedings of the 2023 Conference on Empirical Methods in Natural Language Processing}, pages 530--543. Association for Computational Linguistics, 2023.
\newblock \doi{10.18653/v1/2023.emnlp-main.35}.
\newblock URL \url{https://aclanthology.org/2023.emnlp-main.35/}.

\bibitem[Dai et~al.(2026)Dai, Lin, Xie, and Wang]{dai-etal-2026-realfin}
Yuyang Dai, Yan Lin, Zhuohan Xie, and Yuxia Wang.
\newblock {R}eal{F}in: How well do {LLM}s reason about finance when users leave things unsaid?
\newblock In \emph{Findings of the Association for Computational Linguistics: ACL 2026}, pages 25050--25080, San Diego, California, United States, July 2026. Association for Computational Linguistics.
\newblock \doi{10.18653/v1/2026.findings-acl.1255}.
\newblock URL \url{https://aclanthology.org/2026.findings-acl.1255/}.

\bibitem[Ding et~al.(2026)Ding, Ding, Jiang, Pang, Xiao, Liu, Chen, Zhong, Yuan, Guan, Cheng, and Xu]{ding-etal-2026-cnfinbench}
Jinru Ding, Chao Ding, Yidong Jiang, Wenrao Pang, Boyi Xiao, Zhiqiang Liu, Jiayuan Chen, Yun Zhong, Tiantian Yuan, Junming Guan, Dawei Cheng, and Jie Xu.
\newblock Beyond knowledge to agency: Evaluating expertise, autonomy, and integrity in finance with {CNFinBench}.
\newblock In \emph{Proceedings of the 32nd ACM SIGKDD Conference on Knowledge Discovery and Data Mining V.2}, pages 8765--8776. Association for Computing Machinery, 2026.
\newblock \doi{10.1145/3770855.3817482}.
\newblock URL \url{https://doi.org/10.1145/3770855.3817482}.

\bibitem[Dou et~al.(2026)Dou, Zhu, Wu, Jiang, Li, Guo, Chen, and Zhang]{dou-etal-2026-finguard}
Huaixia Dou, Jie Zhu, Minghao Wu, Shuo Jiang, Junhui Li, Lifan Guo, Feng Chen, and Chi Zhang.
\newblock {F}in{G}uard: Detecting financial regulatory non-compliance in {LLM} interactions.
\newblock arXiv preprint arXiv:2605.29427, 2026.
\newblock URL \url{https://arxiv.org/abs/2605.29427}.

\bibitem[Fei et~al.(2024)Fei, Shen, Zhu, Zhou, Han, Huang, Zhang, Chen, Yin, Shen, Ge, and Ng]{fei-etal-2024-lawbench}
Zhiwei Fei, Xiaoyu Shen, Dawei Zhu, Fengzhe Zhou, Zhuo Han, Alan Huang, Songyang Zhang, Kai Chen, Zhixin Yin, Zongwen Shen, Jidong Ge, and Vincent Ng.
\newblock {L}aw{B}ench: Benchmarking legal knowledge of large language models.
\newblock In \emph{Proceedings of the 2024 Conference on Empirical Methods in Natural Language Processing}, pages 7933--7962. Association for Computational Linguistics, 2024.
\newblock \doi{10.18653/v1/2024.emnlp-main.452}.
\newblock URL \url{https://aclanthology.org/2024.emnlp-main.452/}.

\bibitem[Feng et~al.(2024)Feng, Shi, Wang, Ding, Balachandran, and Tsvetkov]{feng-etal-2024-dont}
Shangbin Feng, Weijia Shi, Yike Wang, Wenxuan Ding, Vidhisha Balachandran, and Yulia Tsvetkov.
\newblock Don{'}t hallucinate, abstain: Identifying {LLM} knowledge gaps via multi-{LLM} collaboration.
\newblock In \emph{Proceedings of the 62nd Annual Meeting of the Association for Computational Linguistics (Volume 1: Long Papers)}, pages 14664--14690. Association for Computational Linguistics, 2024.
\newblock \doi{10.18653/v1/2024.acl-long.786}.
\newblock URL \url{https://aclanthology.org/2024.acl-long.786/}.

\bibitem[Guha et~al.(2023)Guha, Nyarko, Ho, R\'{e}, Chilton, Narayana, Chohlas-Wood, Peters, Waldon, Rockmore, Zambrano, Talisman, Hoque, Surani, Fagan, Sarfaty, Dickinson, Porat, Hegland, Wu, Nudell, Niklaus, Nay, Choi, Tobia, Hagan, Ma, Livermore, Rasumov-Rahe, Holzenberger, Kolt, Henderson, Rehaag, Goel, Gao, Williams, Gandhi, Zur, Iyer, and Li]{guha-etal-2023-legalbench}
Neel Guha, Julian Nyarko, Daniel~E. Ho, Christopher R\'{e}, Adam Chilton, Aditya Narayana, Alex Chohlas-Wood, Austin Peters, Brandon Waldon, Daniel~N. Rockmore, Diego Zambrano, Dmitry Talisman, Enam Hoque, Faiz Surani, Frank Fagan, Galit Sarfaty, Gregory~M. Dickinson, Haggai Porat, Jason Hegland, Jessica Wu, Joe Nudell, Joel Niklaus, John Nay, Jonathan~H. Choi, Kevin Tobia, Margaret Hagan, Megan Ma, Michael Livermore, Nikon Rasumov-Rahe, Nils Holzenberger, Noam Kolt, Peter Henderson, Sean Rehaag, Sharad Goel, Shang Gao, Spencer Williams, Sunny Gandhi, Tom Zur, Varun Iyer, and Zehua Li.
\newblock {LegalBench}: A collaboratively built benchmark for measuring legal reasoning in large language models.
\newblock In \emph{Advances in Neural Information Processing Systems}, volume~36, pages 44123--44279. Curran Associates, Inc., 2023.
\newblock \doi{10.52202/075280-1915}.
\newblock URL \url{https://proceedings.neurips.cc/paper_files/paper/2023/hash/89e44582fd28ddfea1ea4dcb0ebbf4b0-Abstract-Datasets_and_Benchmarks.html}.

\bibitem[Guo et~al.(2025)Guo, Xia, Liu, Cao, Yang, Liu, Wang, Niu, Wang, Wang, Liang, Huang, Zhu, Wei, Chen, Shen, and Zhang]{fineval}
Xin Guo, Haotian Xia, Zhaowei Liu, Hanyang Cao, Zhi Yang, Zhiqiang Liu, Sizhe Wang, Jinyi Niu, Chuqi Wang, Yanhui Wang, Xiaolong Liang, Xiaoming Huang, Bing Zhu, Zhongyu Wei, Yun Chen, Weining Shen, and Liwen Zhang.
\newblock {F}in{E}val: A {C}hinese financial domain knowledge evaluation benchmark for large language models.
\newblock In \emph{Proceedings of the 2025 Conference of the Nations of the Americas Chapter of the Association for Computational Linguistics: Human Language Technologies (Volume 1: Long Papers)}, pages 6258--6292. Association for Computational Linguistics, 2025.
\newblock \doi{10.18653/v1/2025.naacl-long.318}.
\newblock URL \url{https://aclanthology.org/2025.naacl-long.318/}.

\bibitem[He et~al.(2026)He, Gu, Hu, Zhou, Shen, Liang, Dai, Ma, Yu, Xiao, and Li]{he-etal-2026-large}
Ying He, Zhouhong Gu, Zhecheng Hu, Yubo Zhou, Hao Shen, Jiaqing Liang, Zhaoqian Dai, Shuguang Ma, Fei Yu, Yanghua Xiao, and Zhixu Li.
\newblock Are large language models reliable reviewers? a benchmark for error detection in financial documents.
\newblock In \emph{Findings of the Association for Computational Linguistics: ACL 2026}, pages 29625--29643. Association for Computational Linguistics, 2026.
\newblock \doi{10.18653/v1/2026.findings-acl.1481}.
\newblock URL \url{https://aclanthology.org/2026.findings-acl.1481/}.

\bibitem[Huang et~al.(2023)Huang, Bai, Zhu, Zhang, Zhang, Su, Liu, Lv, Zhang, Lei, Fu, Sun, and He]{huang2023ceval}
Yuzhen Huang, Yuzhuo Bai, Zhihao Zhu, Junlei Zhang, Jinghan Zhang, Tangjun Su, Junteng Liu, Chuancheng Lv, Yikai Zhang, Jiayi Lei, Yao Fu, Maosong Sun, and Junxian He.
\newblock {C-Eval}: A multi-level multi-discipline chinese evaluation suite for foundation models.
\newblock In \emph{Advances in Neural Information Processing Systems}, volume~36, pages 62991--63010. Curran Associates, Inc., 2023.
\newblock \doi{10.52202/075280-2749}.
\newblock URL \url{https://proceedings.neurips.cc/paper_files/paper/2023/hash/c6ec1844bec96d6d32ae95ae694e23d8-Abstract-Datasets_and_Benchmarks.html}.

\bibitem[Kim et~al.(2026)Kim, Park, Kim, Jeong, Song, Lim, and Kim]{kim-etal-2026-finred}
Chaeyun Kim, Dae-Young Park, Junghwan Kim, Jinyoung Jeong, Eunji Song, YongTaek Lim, and Minwoo Kim.
\newblock {F}in{RED}: An expert-guided benchmark generation and evaluation framework for financial {LLM} red-teaming.
\newblock arXiv preprint arXiv:2606.19887, 2026.
\newblock URL \url{https://arxiv.org/abs/2606.19887}.

\bibitem[Krumdick et~al.(2024)Krumdick, Koncel-Kedziorski, Lai, Reddy, Lovering, and Tanner]{krumdick-etal-2024-bizbench}
Michael Krumdick, Rik Koncel-Kedziorski, Viet~Dac Lai, Varshini Reddy, Charles Lovering, and Chris Tanner.
\newblock {B}iz{B}ench: A quantitative reasoning benchmark for business and finance.
\newblock In \emph{Proceedings of the 62nd Annual Meeting of the Association for Computational Linguistics (Volume 1: Long Papers)}, pages 8309--8332. Association for Computational Linguistics, 2024.
\newblock \doi{10.18653/v1/2024.acl-long.452}.
\newblock URL \url{https://aclanthology.org/2024.acl-long.452/}.

\bibitem[Li et~al.(2025)Li, Kim, and Wang]{li-etal-2025-questbench}
Belinda~Z. Li, Been Kim, and Zi~Wang.
\newblock {QuestBench}: Can {LLM}s ask the right question to acquire information in reasoning tasks?
\newblock In \emph{Advances in Neural Information Processing Systems}, volume~38. Curran Associates, Inc., 2025.
\newblock \doi{10.52202/085713-4503}.
\newblock URL \url{https://proceedings.neurips.cc/paper_files/paper/2025/hash/c42c8d51556fabb4b57fc86d3d3d0d09-Abstract-Datasets_and_Benchmarks_Track.html}.

\bibitem[Li et~al.(2024)Li, Balachandran, Feng, Ilgen, Pierson, Koh, and Tsvetkov]{li-etal-2024-mediq}
Shuyue~Stella Li, Vidhisha Balachandran, Shangbin Feng, Jonathan~S. Ilgen, Emma Pierson, Pang~Wei Koh, and Yulia Tsvetkov.
\newblock {MediQ}: Question-asking {LLM}s and a benchmark for reliable interactive clinical reasoning.
\newblock In \emph{Advances in Neural Information Processing Systems}, volume~37, pages 28858--28888. Curran Associates, Inc., 2024.
\newblock \doi{10.52202/079017-0908}.
\newblock URL \url{https://proceedings.neurips.cc/paper_files/paper/2024/hash/32b80425554e081204e5988ab1c97e9a-Abstract-Conference.html}.

\bibitem[Liang et~al.(2023)Liang, Bommasani, Lee, Tsipras, Soylu, Yasunaga, Zhang, Narayanan, Wu, Kumar, Newman, Yuan, Yan, Zhang, Cosgrove, Manning, R\'{e}, Acosta-Navas, Hudson, Zelikman, Durmus, Ladhak, Rong, Ren, Yao, Wang, Santhanam, Orr, Zheng, Yuksekgonul, Suzgun, Kim, Guha, Chatterji, Khattab, Henderson, Huang, Chi, Xie, Santurkar, Ganguli, Hashimoto, Icard, Zhang, Chaudhary, Wang, Li, Mai, Zhang, and Koreeda]{liang-etal-2023-holistic}
Percy Liang, Rishi Bommasani, Tony Lee, Dimitris Tsipras, Dilara Soylu, Michihiro Yasunaga, Yian Zhang, Deepak Narayanan, Yuhuai Wu, Ananya Kumar, Benjamin Newman, Binhang Yuan, Bobby Yan, Ce~Zhang, Christian~Alexander Cosgrove, Christopher~D. Manning, Christopher R\'{e}, Diana Acosta-Navas, Drew~Arad Hudson, Eric Zelikman, Esin Durmus, Faisal Ladhak, Frieda Rong, Hongyu Ren, Huaxiu Yao, Jue Wang, Keshav Santhanam, Laurel Orr, Lucia Zheng, Mert Yuksekgonul, Mirac Suzgun, Nathan Kim, Neel Guha, Niladri~S. Chatterji, Omar Khattab, Peter Henderson, Qian Huang, Ryan~Andrew Chi, Sang~Michael Xie, Shibani Santurkar, Surya Ganguli, Tatsunori Hashimoto, Thomas Icard, Tianyi Zhang, Vishrav Chaudhary, William Wang, Xuechen Li, Yifan Mai, Yuhui Zhang, and Yuta Koreeda.
\newblock Holistic evaluation of language models.
\newblock \emph{Transactions on Machine Learning Research}, 2023.
\newblock ISSN 2835-8856.
\newblock URL \url{https://openreview.net/forum?id=iO4LZibEqW}.

\bibitem[Liu et~al.(2023)Liu, Iter, Xu, Wang, Xu, and Zhu]{liu-etal-2023-geval}
Yang Liu, Dan Iter, Yichong Xu, Shuohang Wang, Ruochen Xu, and Chenguang Zhu.
\newblock {G}-eval: {NLG} evaluation using {GPT}-4 with better human alignment.
\newblock In \emph{Proceedings of the 2023 Conference on Empirical Methods in Natural Language Processing}, pages 2511--2522. Association for Computational Linguistics, 2023.
\newblock \doi{10.18653/v1/2023.emnlp-main.153}.
\newblock URL \url{https://aclanthology.org/2023.emnlp-main.153/}.

\bibitem[Matlin et~al.(2025)Matlin, Okamoto, Pardawala, Yang, and Chava]{matlin-etal-2025-financial}
Glenn Matlin, Mika Okamoto, Huzaifa Pardawala, Yang Yang, and Sudheer Chava.
\newblock Financial language model evaluation ({FL}a{ME}).
\newblock In \emph{Findings of the Association for Computational Linguistics: ACL 2025}, pages 22633--22679. Association for Computational Linguistics, 2025.
\newblock \doi{10.18653/v1/2025.findings-acl.1164}.
\newblock URL \url{https://aclanthology.org/2025.findings-acl.1164/}.

\bibitem[Mozannar and Sontag(2020)]{mozannar-sontag-2020-defer}
Hussein Mozannar and David Sontag.
\newblock Consistent estimators for learning to defer to an expert.
\newblock In \emph{Proceedings of the 37th International Conference on Machine Learning}, volume 119 of \emph{Proceedings of Machine Learning Research}, pages 7076--7087. PMLR, 2020.
\newblock URL \url{https://proceedings.mlr.press/v119/mozannar20b.html}.

\bibitem[Nie et~al.(2025)Nie, Yan, Guo, Liu, Wang, He, Zheng, Wang, Li, Sun, Wang, and Tao]{nie-etal-2025-cfinbench}
Ying Nie, Binwei Yan, Tianyu Guo, Hao Liu, Haoyu Wang, Wei He, Binfan Zheng, Weihao Wang, Qiang Li, Weijian Sun, Yunhe Wang, and Dacheng Tao.
\newblock {CF}in{B}ench: A comprehensive {C}hinese financial benchmark for large language models.
\newblock In \emph{Proceedings of the 2025 Conference of the Nations of the Americas Chapter of the Association for Computational Linguistics: Human Language Technologies (Volume 1: Long Papers)}, pages 876--891. Association for Computational Linguistics, 2025.
\newblock \doi{10.18653/v1/2025.naacl-long.40}.
\newblock URL \url{https://aclanthology.org/2025.naacl-long.40/}.

\bibitem[Panickssery et~al.(2024)Panickssery, Bowman, and Feng]{panickssery-etal-2024-llm-evaluators}
Arjun Panickssery, Samuel~R. Bowman, and Shi Feng.
\newblock {LLM} evaluators recognize and favor their own generations.
\newblock In \emph{Advances in Neural Information Processing Systems}, volume~37, pages 68772--68802. Curran Associates, Inc., 2024.
\newblock \doi{10.52202/079017-2197}.
\newblock URL \url{https://proceedings.neurips.cc/paper_files/paper/2024/hash/7f1f0218e45f5414c79c0679633e47bc-Abstract-Conference.html}.

\bibitem[Rao and Daum\'{e}~III(2018)]{rao-daume-iii-2018-learning}
Sudha Rao and Hal Daum\'{e}~III.
\newblock Learning to ask good questions: Ranking clarification questions using neural expected value of perfect information.
\newblock In \emph{Proceedings of the 56th Annual Meeting of the Association for Computational Linguistics (Volume 1: Long Papers)}, pages 2737--2746. Association for Computational Linguistics, 2018.
\newblock \doi{10.18653/v1/P18-1255}.
\newblock URL \url{https://aclanthology.org/P18-1255/}.

\bibitem[Ren et~al.(2023)Ren, Dixit, Bodrova, Singh, Tu, Brown, Xu, Takayama, Xia, Varley, Xu, Sadigh, Zeng, and Majumdar]{ren-etal-2023-ask-help}
Allen~Z. Ren, Anushri Dixit, Alexandra Bodrova, Sumeet Singh, Stephen Tu, Noah Brown, Peng Xu, Leila Takayama, Fei Xia, Jake Varley, Zhenjia Xu, Dorsa Sadigh, Andy Zeng, and Anirudha Majumdar.
\newblock Robots that ask for help: Uncertainty alignment for large language model planners.
\newblock In \emph{Proceedings of the 7th Conference on Robot Learning}, volume 229 of \emph{Proceedings of Machine Learning Research}, pages 661--682. PMLR, 2023.
\newblock URL \url{https://proceedings.mlr.press/v229/ren23a.html}.

\bibitem[Ribeiro et~al.(2020)Ribeiro, Wu, Guestrin, and Singh]{ribeiro-etal-2020-beyond}
Marco~Tulio Ribeiro, Tongshuang Wu, Carlos Guestrin, and Sameer Singh.
\newblock Beyond accuracy: Behavioral testing of {NLP} models with {C}heck{L}ist.
\newblock In \emph{Proceedings of the 58th Annual Meeting of the Association for Computational Linguistics}, pages 4902--4912. Association for Computational Linguistics, 2020.
\newblock \doi{10.18653/v1/2020.acl-main.442}.
\newblock URL \url{https://aclanthology.org/2020.acl-main.442/}.

\bibitem[Shah et~al.(2022)Shah, Chawla, Eidnani, Shah, Du, Chava, Raman, Smiley, Chen, and Yang]{flue}
Raj Shah, Kunal Chawla, Dheeraj Eidnani, Agam Shah, Wendi Du, Sudheer Chava, Natraj Raman, Charese Smiley, Jiaao Chen, and Diyi Yang.
\newblock When {FLUE} meets {FLANG}: Benchmarks and large pretrained language model for financial domain.
\newblock In \emph{Proceedings of the 2022 Conference on Empirical Methods in Natural Language Processing}, pages 2322--2335. Association for Computational Linguistics, 2022.
\newblock \doi{10.18653/v1/2022.emnlp-main.148}.
\newblock URL \url{https://aclanthology.org/2022.emnlp-main.148/}.

\bibitem[Tang et~al.(2025)Tang, E, Ma, He, Liu, Yang, Rong, Li, Ji, Huang, Hu, Liu, and Zheng]{tang-etal-2025-financereasoning}
Zichen Tang, Haihong E, Ziyan Ma, Haoyang He, Jiacheng Liu, Zhongjun Yang, Zihua Rong, Rongjin Li, Kun Ji, Qing Huang, Xinyang Hu, Yang Liu, and Qianhe Zheng.
\newblock {F}inance{R}easoning: Benchmarking financial numerical reasoning more credible, comprehensive and challenging.
\newblock In \emph{Proceedings of the 63rd Annual Meeting of the Association for Computational Linguistics (Volume 1: Long Papers)}, pages 15721--15749. Association for Computational Linguistics, 2025.
\newblock \doi{10.18653/v1/2025.acl-long.766}.
\newblock URL \url{https://aclanthology.org/2025.acl-long.766/}.

\bibitem[Wei et~al.(2026)Wei, Chen, Wang, Huang, Chen, Pan, Li, and Ding]{wei-etal-2026-grounded}
Fei Wei, Daoyuan Chen, Ce~Wang, Yilun Huang, Yushuo Chen, Xuchen Pan, Yaliang Li, and Bolin Ding.
\newblock Grounded in reality: Learning and deploying proactive {LLM} from offline logs.
\newblock In \emph{Proceedings of the 43rd International Conference on Machine Learning}, 2026.
\newblock URL \url{https://openreview.net/forum?id=J4k8q8nku1}.
\newblock ICML 2026 Poster; arXiv:2510.25441.

\bibitem[Xie et~al.(2023)Xie, Han, Zhang, Lai, Peng, Lopez-Lira, and Huang]{xie-etal-2023-pixiu}
Qianqian Xie, Weiguang Han, Xiao Zhang, Yanzhao Lai, Min Peng, Alejandro Lopez-Lira, and Jimin Huang.
\newblock {PIXIU}: A comprehensive benchmark, instruction dataset and large language model for finance.
\newblock In \emph{Advances in Neural Information Processing Systems}, volume~36, pages 33469--33484. Curran Associates, Inc., 2023.
\newblock \doi{10.52202/075280-1454}.
\newblock URL \url{https://proceedings.neurips.cc/paper_files/paper/2023/hash/6a386d703b50f1cf1f61ab02a15967bb-Abstract-Datasets_and_Benchmarks.html}.

\bibitem[Xie et~al.(2024)Xie, Han, Chen, Xiang, Zhang, He, Xiao, Li, Dai, Feng, Xu, Kang, Kuang, Yuan, Yang, Luo, Zhang, Liu, Xiong, Deng, Jiang, Yao, Li, Yu, Hu, Huang, Liu, Lopez-Lira, Wang, Lai, Wang, Peng, Ananiadou, and Huang]{xie2024holisticfinben}
Qianqian Xie, Weiguang Han, Zhengyu Chen, Ruoyu Xiang, Xiao Zhang, Yueru He, Mengxi Xiao, Dong Li, Yongfu Dai, Duanyu Feng, Yijing Xu, Haoqiang Kang, Ziyan Kuang, Chenhan Yuan, Kailai Yang, Zheheng Luo, Tianlin Zhang, Zhiwei Liu, Guojun Xiong, Zhiyang Deng, Yuechen Jiang, Zhiyuan Yao, Haohang Li, Yangyang Yu, Gang Hu, Jiajia Huang, Xiao-Yang Liu, Alejandro Lopez-Lira, Benyou Wang, Yanzhao Lai, Hao Wang, Min Peng, Sophia Ananiadou, and Jimin Huang.
\newblock {FinBen}: A holistic financial benchmark for large language models.
\newblock In \emph{Advances in Neural Information Processing Systems}, volume~37, pages 95716--95743. Curran Associates, Inc., 2024.
\newblock \doi{10.52202/079017-3033}.
\newblock URL \url{https://proceedings.neurips.cc/paper_files/paper/2024/hash/adb1d9fa8be4576d28703b396b82ba1b-Abstract-Datasets_and_Benchmarks_Track.html}.

\bibitem[Zhang et~al.(2024)Zhang, Qin, Deng, Huang, Lei, Liu, Jin, Liang, and Chua]{zhang-etal-2024-clamber}
Tong Zhang, Peixin Qin, Yang Deng, Chen Huang, Wenqiang Lei, Junhong Liu, Dingnan Jin, Hongru Liang, and Tat-Seng Chua.
\newblock {CLAMBER}: A benchmark of identifying and clarifying ambiguous information needs in large language models.
\newblock In \emph{Proceedings of the 62nd Annual Meeting of the Association for Computational Linguistics (Volume 1: Long Papers)}, pages 10746--10766. Association for Computational Linguistics, 2024.
\newblock \doi{10.18653/v1/2024.acl-long.578}.
\newblock URL \url{https://aclanthology.org/2024.acl-long.578/}.

\bibitem[Zhao et~al.(2026)Zhao, Fang, and Cheng]{zhao-etal-2026-ask}
Jiale Zhao, Ke~Fang, and Lu~Cheng.
\newblock When and what to ask: {AskBench} and rubric-guided {RLVR} for {LLM} clarification.
\newblock In \emph{Findings of the Association for Computational Linguistics: ACL 2026}, pages 17120--17140. Association for Computational Linguistics, 2026.
\newblock \doi{10.18653/v1/2026.findings-acl.845}.
\newblock URL \url{https://aclanthology.org/2026.findings-acl.845/}.

\bibitem[Zhao et~al.(2024)Zhao, Gao, Cardie, and Rush]{zhao-etal-2024-couldve}
Wenting Zhao, Ge~Gao, Claire Cardie, and Alexander~M. Rush.
\newblock {I} could{'}ve asked that: Reformulating unanswerable questions.
\newblock In \emph{Proceedings of the 2024 Conference on Empirical Methods in Natural Language Processing}, pages 4207--4220. Association for Computational Linguistics, 2024.
\newblock \doi{10.18653/v1/2024.emnlp-main.242}.
\newblock URL \url{https://aclanthology.org/2024.emnlp-main.242/}.

\bibitem[Zheng et~al.(2023)Zheng, Chiang, Sheng, Zhuang, Wu, Zhuang, Lin, Li, Li, Xing, Zhang, Gonzalez, and Stoica]{zheng-etal-2023-judging}
Lianmin Zheng, Wei-Lin Chiang, Ying Sheng, Siyuan Zhuang, Zhanghao Wu, Yonghao Zhuang, Zi~Lin, Zhuohan Li, Dacheng Li, Eric~P. Xing, Hao Zhang, Joseph~E. Gonzalez, and Ion Stoica.
\newblock Judging {LLM}-as-a-judge with {MT}-bench and chatbot arena.
\newblock In \emph{Advances in Neural Information Processing Systems}, volume~36, pages 46595--46623. Curran Associates, Inc., 2023.
\newblock \doi{10.52202/075280-2020}.
\newblock URL \url{https://proceedings.neurips.cc/paper_files/paper/2023/hash/91f18a1287b398d378ef22505bf41832-Abstract-Datasets_and_Benchmarks.html}.

\bibitem[Zhou et~al.(2025)Zhou, Feng, Zhu, Yao, Koyejo, and Han]{zhou-etal-2025-active}
Zhanke Zhou, Xiao Feng, Zhaocheng Zhu, Jiangchao Yao, Sanmi Koyejo, and Bo~Han.
\newblock From passive to active reasoning: Can large language models ask the right questions under incomplete information?
\newblock In \emph{Proceedings of the 42nd International Conference on Machine Learning}, volume 267 of \emph{Proceedings of Machine Learning Research}, pages 78714--78758. PMLR, 2025.
\newblock URL \url{https://proceedings.mlr.press/v267/zhou25e.html}.

\bibitem[Zhu et~al.(2021)Zhu, Lei, Huang, Wang, Zhang, Lv, Feng, and Chua]{zhu-etal-2021-tat}
Fengbin Zhu, Wenqiang Lei, Youcheng Huang, Chao Wang, Shuo Zhang, Jiancheng Lv, Fuli Feng, and Tat-Seng Chua.
\newblock {TAT}-{QA}: A question answering benchmark on a hybrid of tabular and textual content in finance.
\newblock In \emph{Proceedings of the 59th Annual Meeting of the Association for Computational Linguistics and the 11th International Joint Conference on Natural Language Processing (Volume 1: Long Papers)}, pages 3277--3287. Association for Computational Linguistics, 2021.
\newblock \doi{10.18653/v1/2021.acl-long.254}.
\newblock URL \url{https://aclanthology.org/2021.acl-long.254/}.

\bibitem[Zhu et~al.(2026)Zhu, Huang, Mu, Huang, Nie, Liu, Zhang, Liu, and Zhang]{zhu-etal-2026-diagnosisarena}
Yakun Zhu, Zhongzhen Huang, Linjie Mu, Yutong Huang, Wei Nie, Jiaji Liu, Shaoting Zhang, Pengfei Liu, and Xiaofan Zhang.
\newblock {D}iagnosis{A}rena: Benchmarking diagnostic reasoning for large language models.
\newblock In \emph{Findings of the Association for Computational Linguistics: ACL 2026}, pages 3074--3098, San Diego, California, United States, July 2026. Association for Computational Linguistics.
\newblock \doi{10.18653/v1/2026.findings-acl.151}.
\newblock URL \url{https://aclanthology.org/2026.findings-acl.151/}.

\end{thebibliography}

\newpage
\appendix

\section{Benchmark Construction and Inventory}\label{app:construction}
\label{app:inventory}

    \subsection{Design Comparison with Existing Benchmarks}\label{app:benchmark-comparison}
    
    Table~\ref{tab:benchmark-comparison} compares FinRiskAtlas with representative financial and information-acquisition benchmarks along the design dimensions relevant to our evaluation framework. A property is marked as present only when it serves as a benchmark-level organizing principle rather than an isolated task capability. The comparison is intended to clarify differences in evaluation design, rather than to rank existing benchmarks by overall quality.
    
    \begin{table*}[t]
    \centering
    \footnotesize
    \setlength{\tabcolsep}{3pt}
    \caption{Positioning of FinRiskAtlas relative to representative financial and information-acquisition benchmarks. \emph{Operation-aligned} indicates that downstream units are defined by their role in a professional workflow rather than source format or task type. \emph{Decision contract} indicates that the visible evidence regime, decision object, reviewer artifact, and scoring protocol are fixed at the family level. \emph{Evidence acquisition} indicates evaluation of whether and what additional information should be requested. \ding{51} denotes present as a benchmark-level design unit, \ding{55} not documented as such, and $\circ$ partial.}
    \label{tab:benchmark-comparison}
    \begin{tabularx}{\textwidth}{@{}lYcccc@{}}
    \toprule
    \textbf{Benchmark} & \textbf{Primary focus} & \textbf{\shortstack{Operation\\aligned}} & \textbf{\shortstack{Decision\\contract}} & \textbf{\shortstack{Evidence\\acquisition}} & \textbf{\shortstack{Expert\\trajectories}}\\
    \midrule
    FLUE \cite{flue} & Financial language understanding & \ding{55} & \ding{55} & \ding{55} & \ding{55}\\
    FinQA / ConvFinQA \cite{chen-etal-2021-finqa,chen-etal-2022-convfinqa} & Numerical reasoning over financial reports & \ding{55} & \ding{55} & \ding{55} & \ding{55}\\
    BizBench \cite{krumdick-etal-2024-bizbench} & Business quantitative reasoning & \ding{55} & \ding{55} & \ding{55} & \ding{55}\\
    PIXIU / FinBen \cite{xie-etal-2023-pixiu,xie2024holisticfinben} & Broad financial capability coverage & \ding{55} & $\circ$ & \ding{55} & \ding{55}\\
    CFinBench / FinEval \cite{nie-etal-2025-cfinbench,fineval} & Chinese financial knowledge evaluation & \ding{55} & $\circ$ & \ding{55} & \ding{55}\\
    CNFinBench \cite{ding-etal-2026-cnfinbench} & Financial capability, compliance, and safety & $\circ$ & $\circ$ & \ding{55} & \ding{55}\\
    FinGuard \cite{dou-etal-2026-finguard} & Regulation-derived compliance evaluation & $\circ$ & $\circ$ & \ding{55} & \ding{55}\\
    FinED-Bench \cite{he-etal-2026-large} & Financial document error detection & $\circ$ & \ding{55} & \ding{55} & \ding{55}\\
    Finance Agent \cite{bigeard-etal-2025-financeagent} & Tool-supported financial research agents & $\circ$ & \ding{55} & $\circ$ & \ding{55}\\
    MediQ / QuestBench \cite{li-etal-2024-mediq,li-etal-2025-questbench} & Clarification under incomplete information & \ding{55} & \ding{55} & \ding{51} & \ding{55}\\
    RealFin \cite{dai-etal-2026-realfin} & Unanswerable financial premises & \ding{55} & \ding{55} & $\circ$ & \ding{55}\\
    Learn-to-Ask \cite{wei-etal-2026-grounded} & Policy learning from expert dialogue & \ding{55} & \ding{55} & \ding{51} & \ding{51}\\
    \midrule
    \textbf{FinRiskAtlas} & Operation-level financial risk and compliance review & \ding{51} & \ding{51} & \ding{51} & \ding{51}\\
    \bottomrule
    \end{tabularx}
    \end{table*}
    
    This appendix describes the benchmark inventory, construction process, data sources, and quality-control procedures. The evaluation units and family contracts are defined in Section~\ref{sec:benchmark}, while the FinRisk-Ask reconstruction protocol is described in Appendix~\ref{app:finriskask-protocol}. Static instances and trajectory states represent different evaluation units and are therefore reported separately.
    
    \subsection{Capability-First Construction Pipeline}
    
    FinRiskAtlas follows a capability-first construction process. The central principle is that data collection procedures should instantiate predefined evaluation contracts rather than determine the benchmark organization. We first define the professional capability or review operation, specify its evaluation contract, identify suitable source materials, construct candidate instances, and then perform expert validation.
    
    Three construction routes are used.
    
    \textbf{Direct normalization} converts expert-authored questions and verified answers into standardized evaluation items while preserving their original capability targets and reference decisions.
    
    \textbf{Structured transformation} converts financial and legal records into evidence-processing tasks such as extraction, matching, classification, and quantitative reasoning while preserving the relationship between source evidence and evaluation targets.
    
    \textbf{Source-conditioned construction} derives case-level review tasks from complex business, transaction, or legal materials when the expected output is supported by model-visible evidence, documented professional decisions, or deterministic calculations.
    
    The construction route determines how an item is obtained, but not how it is evaluated. After construction, every candidate item is assigned to a family only when it instantiates the same evaluation contract as other items in that family. This separation prevents source format, collection procedure, or annotation availability from becoming the implicit definition of a benchmark unit.
    
    \begin{table*}[t]
    \centering
    \footnotesize
    \setlength{\tabcolsep}{4pt}
    \caption{Benchmark composition of FinRiskAtlas. Static instances and active-review states represent different evaluation units and are not combined.}
    \label{tab:appendix-inventory}
    \begin{tabularx}{\textwidth}{@{}lXcc@{}}
    \toprule
    \textbf{Layer} & \textbf{Evaluation focus} & \textbf{Families / settings} & \textbf{Instances}\\
    \midrule
    Domain Knowledge & Financial concepts, regulations, obligations, and risk mechanisms supporting professional review & 42 & 4,679\\
    Evidence-Grounded Processing & Transforming heterogeneous records into structured and decision-relevant evidence & 5 & 1,502\\
    Applied Review & Evidence-based case analysis, rule grounding, and professional review generation & 6 & 3,561\\
    \midrule
    Static benchmark & Three-layer financial review evaluation & 53 & 9,742\\
    FinRisk-Ask & Evidence-state control over reconstructed professional review states & 1 setting & 680 states (583 Ask, 97 Proceed)\\
    \bottomrule
    \end{tabularx}
    \end{table*}
    
    Table~\ref{tab:appendix-inventory} summarizes the benchmark composition. The detailed family inventory is provided in Table~\ref{tab:family-inventory}. Family sizes are intentionally unequal because construction follows capability coverage rather than balanced sampling. Aggregation is therefore performed only over explicitly compatible family sets.
    
    \begin{table*}[t]
    \centering
    \footnotesize
    \setlength{\tabcolsep}{4pt}
    \caption{Group-level composition of FinRiskAtlas. Each row reports the taxonomy group, coverage, number of families, and total number of instances. The complete per-family inventory, including sources and contracts, is released with the benchmark package.}
    \label{tab:family-inventory}
    \begin{tabularx}{\textwidth}{@{}llXcc@{}}
    \toprule
    \textbf{Layer} & \textbf{Group} & \textbf{Coverage} & \textbf{Families} & \textbf{Instances}\\
    \midrule
    \multirow{16}{*}{\rotatebox[origin=c]{90}{Domain Knowledge}} & Financial risk & Risk management, markets, products, valuation, credit, operational, liquidity, and fixed-income risk & 8 & 892\\
    & Fraud risk & Card, insurance, telecom, contract, invoice, and AI-enabled fraud & 6 & 624\\
    & Money-laundering risk & Anti-money-laundering obligations and typologies & 2 & 244\\
    & Compliance risk & Regulatory obligations and compliance controls & 4 & 442\\
    & Trade theory and policy & International trade theory and policy & 4 & 457\\
    & FDI and multinationals & Foreign direct investment and multinational operations & 2 & 208\\
    & International finance & Cross-border finance and settlement & 4 & 499\\
    & International business & Business environment and operations & 4 & 438\\
    & FinTech & Financial technology & 1 & 103\\
    & AI security & AI-related security risk & 1 & 106\\
    & Technical security & Security controls & 1 & 103\\
    & Security engineering & Security engineering practice & 1 & 104\\
    & E-commerce & E-commerce risk and operations & 1 & 139\\
    & Payments & Payment systems and controls & 1 & 109\\
    & Finance law & Finance-related legal knowledge & 1 & 110\\
    & Law & General legal knowledge for review & 1 & 101\\
    \cmidrule(l){2-5}
    & \textbf{Subtotal} & & \textbf{42} & \textbf{4,679}\\
    \midrule
    \multirow{4}{*}{\rotatebox[origin=c]{90}{\shortstack{Evidence-\\Grounded}}} & Case classification & Identifying case structure and category & 1 & 445\\
    & Information extraction & Extracting requested fields from source records & 1 & 249\\
    & Entity matching & Institution and person matching & 2 & 750\\
    & Quantitative reasoning & Decision-relevant financial computation & 1 & 58\\
    \cmidrule(l){2-5}
    & \textbf{Subtotal} & & \textbf{5} & \textbf{1,502}\\
    \midrule
    \multirow{4}{*}{\rotatebox[origin=c]{90}{Applied Review}} & Risk classification & Case-level risk categorization & 1 & 366\\
    & Legal decision & Legal-outcome prediction and legal-judgment generation & 2 & 1,445\\
    & Provision grounding & Applicable-provision selection & 1 & 1,000\\
    & Review generation & Decision-view and disputed-issue generation & 2 & 750\\
    \cmidrule(l){2-5}
    & \textbf{Subtotal} & & \textbf{6} & \textbf{3,561}\\
    \midrule
    \multicolumn{3}{@{}l}{\textbf{Static benchmark total}} & \textbf{53} & \textbf{9,742}\\
    \multicolumn{3}{@{}l}{\textbf{FinRisk-Ask active review}} & 1 setting & 680 states\\
    \bottomrule
    \end{tabularx}
    \end{table*}
    
    \subsection{Data Sources and Expert Involvement}
    
    FinRiskAtlas integrates multiple source categories covering different stages of professional financial review. Static benchmark construction involved six domain experts: four with experience in financial risk control and compliance review and two with experience in legal review of financial and commercial disputes. Experts participated in capability definition, family-boundary validation, candidate-item review, ambiguity analysis, and quality control.
    
    \begin{table*}[t]
    \centering
    \footnotesize
    \setlength{\tabcolsep}{4pt}
    \caption{Data sources and expert roles in benchmark construction.}
    \label{tab:data-source-expert}
    \begin{tabularx}{\textwidth}{@{}lXX@{}}
    \toprule
    \textbf{Source category} & \textbf{Capability coverage} & \textbf{Expert involvement}\\
    \midrule
    Professional examinations and textbooks & Financial concepts, regulations, and risk mechanisms & Reviewed capability alignment and reference correctness\\
    Regulatory documents and industry guidance & Compliance obligations and rule-grounded reasoning & Reviewed regulatory interpretation and applicability\\
    Business, transaction, entity, and legal records & Evidence extraction, entity reconciliation, quantitative reasoning, and applied review & Reviewed decision targets and evidence support\\
    Completed risk-review trajectories & Ask-or-Proceed transitions and evidence-request evaluation & Verified reconstructed states, action mappings, and evidence targets\\
    \bottomrule
    \end{tabularx}
    \end{table*}
    
    Every retained instance underwent independent review by at least two experts with relevant financial or legal backgrounds before inclusion. Disagreements were resolved through adjudication, and items without stable expert consensus were removed rather than assigned majority labels. The released benchmark therefore contains consensus references rather than unresolved annotations.
    
    \subsection{Quality Control and Leakage Prevention}
    
    Quality control is conducted at both the family and instance levels.
    
    At the family level, experts verify that all items within a family share the same target capability, model-visible information regime, professional decision object, reviewer artifact, and scoring protocol. Candidate families are split when they contain distinct decision objects and merged when their contracts are equivalent.
    
    At the instance level, we apply the following checks:
    \begin{itemize}[leftmargin=*,nosep]
    \item \textbf{Evidence support:} reference outputs must be supported by supplied evidence, authoritative documents, deterministic transformations, or adjudicated expert decisions.
    \item \textbf{Ambiguity control:} items with unclear decision targets or unstable scoring criteria are revised or removed.
    \item \textbf{Duplicate control:} normalized inputs, source identifiers, and templates are examined to reduce accidental duplication.
    \item \textbf{Leakage control:} answer-bearing metadata, hidden targets, and unavailable future information are removed from model-visible contexts.
    \item \textbf{Parser validation:} released examples and adversarial formatting variants are used to validate deterministic output extraction.
    \item \textbf{Provenance tracking:} every released item is associated with source information and construction records.
    \end{itemize}
    
    For publicly available materials, provenance information is retained and knowledge-oriented evaluation is distinguished from record-grounded review operations, where correctness depends on evidence available within the evaluation context rather than memorized facts. Family-level and instance-level review jointly enforce the five-part evaluation contract: the former defines the measurement unit, while the latter verifies that each retained item is a valid instantiation of that unit.

\section{Evaluation Protocol and Reproducibility}
\label{app:protocol}

    This appendix specifies the execution protocol, response processing, scoring procedures, semantic evaluation, and reproducibility practices of FinRiskAtlas. The benchmark follows a unified evaluation pipeline in which raw model responses, parsed predictions, and final evaluation outputs are retained for verification. FinRiskAtlas is evaluation-only: candidate models are not trained, adapted, or optimized on any benchmark component.
    
    \subsection{Inference Configuration}
    
    All experiments use zero-shot direct-answer inference without in-context demonstrations or requested chain-of-thought. Models receive the task instruction and the model-visible information defined by the corresponding family contract.
    
    All reported results correspond to archived evaluation runs. The run manifest records the exact model identifier, provider or checkpoint revision, decoding configuration, output length limit, evaluation timestamp, retry status, and prediction-file hash for every evaluated configuration. These records define the execution environment required to reproduce the reported results.
    
    Each model--instance pair is evaluated once under the archived configuration. Empty, failed, or unparseable responses remain in the evaluation denominator. Therefore, reported values should be interpreted as point estimates of the evaluated configurations rather than estimates over repeated stochastic generations.
    
    \subsection{Prompt and Output Processing}
    
    During evaluation, each family contract is instantiated through a fixed prompt template, output schema, parser, and scoring implementation. The output contract determines the expected response format and the corresponding parsing procedure before candidate evaluation.
    
    Knowledge and classification tasks require labels, options, or option sets. Information-extraction tasks require structured fields. Entity-reconciliation tasks require identity decisions. Quantitative reasoning tasks require normalized numerical answers. Open-ended Applied Review tasks require professional analyses evaluated by semantic rubrics. FinRisk-Ask requires either an Ask decision with one evidence request or a Proceed decision.
    
    The evaluation pipeline preserves:
    
    \begin{enumerate}[leftmargin=*,nosep]
        \item raw model responses before parsing;
        \item parsed outputs produced by deterministic parsers;
        \item final metric values produced by family-specific scoring functions.
    \end{enumerate}
    
    Answer normalization is restricted to formatting variation and does not modify the underlying prediction content. Parsed outputs are retained so that parsing success and valid-only analyses can be recovered without changing the primary evaluation protocol.
    
    \subsection{Task-Native Scoring Protocol}
    
    For a model configuration $m$ and a task family $f$ containing $N_f$ instances, let $\hat{y}_{m,i}$ denote the raw response to instance $i$, let $p_f$ denote the fixed family parser, and let $g_f\in[0,1]$ denote the family-specific scoring function. The configuration--family score is
    \begin{equation}
    S_{m,f}
    =
    \frac{100}{N_f}
    \sum_{i=1}^{N_f}
    g_f\!\left(p_f(\hat{y}_{m,i}),y_i\right).
    \label{eq:family-score}
    \end{equation}
    Malformed or unparseable outputs are mapped by $p_f$ to a designated invalid value and receive zero credit under $g_f$. For a downstream operation $o$, we write $S_{m,o}$ for the corresponding operation-family score.
    
    \begin{table*}[t]
        \centering
        \scriptsize
        \setlength{\tabcolsep}{3.5pt}
        \caption{Task-native evaluation contracts in FinRiskAtlas.}
        \label{tab:scoring-contracts}
        \begin{tabularx}{\textwidth}{@{}l l X l@{}}
        \toprule
        \textbf{Task type} & \textbf{Output} & \textbf{Scoring rule} & \textbf{Metric}\\
        \midrule
        Knowledge and classification & Label, option, or option set & Correctness after family-specific extraction, set handling, and normalization & Accuracy-style score\\
        Information extraction & Requested fields & Normalized field-level comparison with reference fields, preserving required field identity and order & Field-level score\\
        Entity reconciliation & Entity identity & Exact correspondence with the reference identity decision & Accuracy\\
        Quantitative reasoning & Numerical answer & Family-specific numerical normalization and correctness checking & Accuracy-style score\\
        Risk and legal decisions & Categorical decision & Exact decision matching after label normalization & Accuracy\\
        Open-ended review generation & Professional analysis & Fixed semantic rubric evaluating correctness, coverage, and evidence use & Rubric score\\
        FinRisk-Ask action & Ask or Proceed & Agreement with the recorded reviewer transition & RAA, AskR, ProceedR, BAcc\\
        FinRisk-Ask evidence request & Focused evidence request & Three-level semantic alignment with later-observed, expert-verified evidence targets & ERA, CRA, DEH, REH\\
        \bottomrule
        \end{tabularx}
    \end{table*}

    Table~\ref{tab:operation-contracts} instantiates the evaluation contract $\Gamma_f=(c_f,\mathcal{I}_f,d_f,\mathcal{Y}_f,s_f)$ for each of the eleven fixed-evidence downstream operations. The operation name specifies $c_f$, while the remaining columns summarize $\mathcal{I}_f$, $d_f$, $\mathcal{Y}_f$, and $s_f$.
    
    \begin{table*}[t]
        \centering
        \scriptsize
        \setlength{\tabcolsep}{3.2pt}
        \renewcommand{\arraystretch}{1.08}
        \caption{\textbf{Evaluation contracts for the eleven fixed-evidence downstream operations.} The operation name specifies the target capability $c_f$; the remaining columns summarize the model-visible information regime $\mathcal{I}_f$, decision object $d_f$, reviewer-artifact schema $\mathcal{Y}_f$, and scoring protocol $s_f$.}
        \label{tab:operation-contracts}
        \begin{tabularx}{\textwidth}{@{}p{0.15\textwidth}YYYY@{}}
        \toprule
        \textbf{Operation} & \textbf{Model-visible information} & \textbf{Decision object} & \textbf{Reviewer artifact} & \textbf{Scoring}\\
        \midrule
        \multicolumn{5}{@{}l}{\textit{Evidence-Grounded Processing}}\\[-1pt]
        Case classification & Full case record and a fixed candidate category set & Case category & Procedure-routing label & Accuracy\\
        Information extraction & Source record and an ordered list of requested fields & Values of the requested fields & Structured evidence-field sequence & Normalized field-level score\\
        Institution matching & Two institution descriptions, possibly from different records or languages & Whether the descriptions denote the same institution & Institution-identity decision & Accuracy\\
        Person matching & Two person records with partially overlapping attributes & Whether the records denote the same natural person & Person-identity decision & Accuracy\\
        Quantitative reasoning & Case figures together with the applicable definition or formula & Requested numerical quantity & One normalized quantity & Numerical correctness\\
        \addlinespace[3pt]
        \multicolumn{5}{@{}l}{\textit{Applied Review}}\\[-1pt]
        Risk classification & Case-level evidence and the applicable risk-category set & Case-level risk category & Risk grade & Accuracy\\
        Applicable-provision selection & Case facts and a candidate provision set & Governing provisions & Provision set cited as the decision basis & Set-based accuracy-style score\\
        Legal-outcome prediction & Case facts and procedural record, with the later outcome withheld & Case disposition & Disposition label & Accuracy\\
        Legal-judgment generation & Case facts, evidence, and procedural record & Reasoned analysis supporting a disposition & Professional legal analysis & Fixed semantic rubric\\
        Disputed-issue generation & Case record containing overlapping legal and contractual relationships & Set of contested issues & Structured issue list & Fixed semantic rubric\\
        Decision-view generation & Case record and the review question presented to the reviewer & Recommended review position & Professional review opinion & Fixed semantic rubric\\
        \bottomrule
        \end{tabularx}
    \end{table*}
    
    Let $\mathcal{F}_{K}$ denote the set of 42 Domain Knowledge families. The Domain Knowledge macro score of configuration $m$ is
    \begin{equation}
    K_m
    =
    \frac{1}{|\mathcal{F}_{K}|}
    \sum_{f\in\mathcal{F}_{K}} S_{m,f},
    \qquad
    |\mathcal{F}_{K}|=42.
    \label{eq:knowledge-macro}
    \end{equation}
    Each knowledge family therefore contributes equally regardless of its number of instances. No corresponding cross-operation macro is used as a primary result because downstream operations produce different reviewer artifacts and use different native scoring protocols.
    
    \subsection{Semantic Evaluation Protocol}
    
    Three open-ended Applied Review operations use semantic evaluation: legal-judgment generation, decision-view generation, and disputed-issue generation. Candidate responses are evaluated by a fixed semantic evaluator, DeepSeek-V4-Flash, following established LLM-based, rubric-guided evaluation practice for open-ended generation~\cite{liu-etal-2023-geval,zheng-etal-2023-judging}.
    
    The evaluator receives:
    
    \begin{itemize}[leftmargin=*,nosep]
        \item the original task input;
        \item reference information;
        \item the candidate response;
        \item the task-specific evaluation rubric.
    \end{itemize}
    
    Candidate-model identity is not provided as an evaluator input. Evaluator configuration, prompts, and parsing procedures are fixed independently of candidate generation and applied uniformly to all responses.
    
    The evaluator configuration's own candidate outputs are retained for transparency but excluded from best-value annotations, operation-level comparisons, and analyses involving judge-scored tasks.
    
    FinRisk-Ask request alignment uses the same fixed evaluator with a three-level rubric. For each retained Ask state, the evaluator receives the generated request and the expert-verified future evidence target set. A score of 1 indicates direct alignment, 0.5 indicates a relevant but incomplete request, and 0 indicates an unrelated request. The evaluator does not determine the Ask-or-Proceed reference action, which is derived from the recorded workflow transition.
    
    The semantic evaluator is treated as a fixed measurement instrument rather than an oracle of professional correctness. Excluding evaluator self-comparison and releasing evaluator artifacts improve reproducibility but do not eliminate evaluator-specific effects~\cite{zheng-etal-2023-judging,panickssery-etal-2024-llm-evaluators}. Semantic scores should therefore be interpreted as rubric measurements under the released evaluator configuration.
    
    \subsection{Aggregation and Statistical Interpretation}
    
    Reported results are descriptive comparisons among archived configurations. Several configurations belong to related model families and should not be interpreted as independent samples from a model population. FinRisk-Ask states originate from 104 trajectories, and multiple states from the same trajectory may share case-level dependencies.
    
    Interpretation of reported comparisons follows several principles. First, analyses involving operation maxima refer to observed point-estimate maxima rather than universal model rankings. Second, regret analyses consider near-tied Domain Knowledge configurations to reduce sensitivity to arbitrary ordering. Third, Ask-or-Proceed comparisons are evaluated against one shared recorded workflow boundary rather than treated as independent hypothesis tests. Fourth, FinRisk-Ask metrics are state-weighted, meaning trajectories contributing more retained states receive greater influence.
    
    Percentile profiles are used only for comparing relative rankings across heterogeneous operations. They do not imply that extraction, classification, numerical reasoning, and rubric-based generation share a common performance scale.
    
    \subsection{Reproducibility Artifacts}
    
    The released evaluation package includes the benchmark manifest, prompts, inference parameters, raw and parsed predictions, evaluator configurations, scoring implementations, and analysis scripts. These artifacts allow new model configurations to be evaluated under the same contracts and compared on the same operation and evidence-state dimensions.

\section{FinRisk-Ask Protocol}\label{app:finriskask-protocol}

    FinRisk-Ask is an offline trajectory-replay evaluation of evidence-state control in financial review. Unlike interactive information-seeking benchmarks and policy-learning approaches~\cite{li-etal-2024-mediq,li-etal-2025-questbench,zhou-etal-2025-active,zhao-etal-2026-ask,wei-etal-2026-grounded}, it does not optimize a dialogue policy or evaluate long-horizon interaction. Instead, it measures whether a model can make an appropriate evidence-state transition under a fixed historical decision boundary and whether a generated request addresses a trajectory-supported evidence need.
    
    \subsection{Trajectory Reconstruction and Release Filtering}
    
    FinRisk-Ask is constructed from completed financial-risk review trajectories collected from an internal enterprise risk-control workflow. The trajectories originate from real review processes and are released only through de-identified reconstructed states.
    
    For each trajectory, reconstruction proceeds chronologically to identify decision points where the recorded workflow either initiates additional evidence acquisition or advances using the currently available record. A candidate state is retained when three conditions are satisfied: (i) the pre-action context can be reconstructed without later observations, (ii) the immediately following reviewer action maps unambiguously to Ask or Proceed, and (iii) no post-action information is included in the model-visible context.
    
    Ask-labeled states require one additional condition. At least one later-observed evidence item must be available from the completed trajectory and verified by experts as unavailable at the decision point, relevant to an unresolved review need, and appropriate as an evidence-acquisition target. During reconstruction, 10 Ask-labeled states were removed because no admissible future evidence target could be identified and verified. The released benchmark therefore contains 680 states from 104 trajectories, including 583 recorded Ask states and 97 recorded Proceed states.
    
    The filtering rule is defined independently of candidate-model outputs. Every released state supports action evaluation, and every released Ask state additionally supports request evaluation, providing a shared denominator for AskR, ERA, and CRA. Multiple states may originate from the same trajectory, but each state preserves the evidence boundary that existed at its own historical decision point.
    
    For each released decision point $t$, $C_t$ denotes the complete case evidence and review history available immediately before the recorded action $a_t^*$. Subsequent evidence, later reviewer actions, and downstream outcomes are excluded from $C_t$ and are accessible only to the evaluation pipeline.
    
    \subsection{Reference Action and Evidence Target Construction}
    
    The Ask-or-Proceed reference is derived from the recorded reviewer action immediately following each reconstructed state. It is not inferred from whether additional evidence appears later in the trajectory. A state is labeled Ask when the recorded action initiates a new evidence-acquisition request, and Proceed when the workflow advances using the currently available record without initiating a request at that point. Evidence acquisition that occurs at later decision points does not change the current label.
    
    The reference action is defined as:
    \begin{equation}
    a_t^*=
    \begin{cases}
    \mathrm{Ask}, & \text{if the recorded action at }t\text{ initiates additional evidence acquisition},\\
    \mathrm{Proceed}, & \text{if the workflow advances at }t\text{ without initiating a new request}.
    \end{cases}
    \end{equation}
    The released labels contain 583 Ask states and 97 Proceed states. Action mappings were reviewed by domain experts, and disagreements were resolved through adjudication. The reference represents an observed professional transition under one workflow rather than a claim that it is the only defensible review policy.
    
    For retained Ask states, FinRisk-Ask constructs an expert-verified set of later-observed evidence needs:
    \begin{equation}
    \mathcal{E}_t^{+}=\{e_{t,1},e_{t,2},\ldots,e_{t,n_t}\},\qquad n_t\geq 1.
    \end{equation}
    Historical reviewer requests are not used as exact textual references. Instead, generated requests are evaluated against the underlying evidence need, allowing semantically equivalent requests to receive credit.
    
    Future evidence targets are constructed by collecting later-observed materials after the decision point, normalizing evidence descriptions, and verifying the resulting targets with domain experts. A target is retained only if experts confirm that the evidence is temporally unavailable at state $t$, addresses an unresolved review need, and corresponds to a meaningful evidence-acquisition objective.
    
    Candidate models never observe $\mathcal{E}_t^{+}$ or any other future trajectory information during inference.
    
    \subsection{Expert Validation and Quality Control}
    
    FinRisk-Ask applies expert validation to three aspects of trajectory reconstruction: state validity, action mapping, and evidence-target admissibility. Experts verify that reconstructed states preserve the original information boundary, that Ask-or-Proceed labels correspond to the recorded workflow transition, and that retained evidence targets represent unresolved decision-relevant needs at the replayed state.
    
    The quality-control process separates three requirements:
    \begin{itemize}[leftmargin=*,nosep]
    \item \textbf{Temporal validity:} target evidence must appear only after the evaluated decision point and must not already exist in the visible context.
    \item \textbf{Decision relevance:} the evidence must address a need that affects the professional decision at that state.
    \item \textbf{Semantic validity:} different descriptions referring to the same evidence need should be recognized as equivalent.
    \end{itemize}
    Targets failing any of these requirements are removed or revised before release.
    
    \subsection{Evaluation Protocol and Parsing}
    
    Given the pre-action context $C_t$, the required structured model output is
    \begin{equation}
    \hat{y}_t=
    \begin{cases}
    \mathrm{Ask}(\hat{q}_t), & \text{request one focused item of additional evidence},\\
    \mathrm{Proceed}, & \text{continue with the currently available evidence}.
    \end{cases}
    \label{eq:finriskask-output}
    \end{equation}
    The action parser maps $\hat{y}_t$ to $\hat{a}_t\in\{\mathrm{Ask},\mathrm{Proceed},\mathrm{Invalid}\}$. When $\hat{a}_t=\mathrm{Ask}$, the request parser additionally extracts $\hat{q}_t$ and checks whether it satisfies the one-request output contract. A parsed Ask action with an empty, malformed, rejected, or multi-item request remains an Ask prediction for AskR but receives zero request-alignment credit. A structurally valid but semantically unrelated request likewise receives zero credit under the alignment rubric. An Invalid action receives zero action-agreement credit, while a Proceed prediction on a recorded Ask state receives zero ERA credit because no request is produced.
    
    The parsing rules and normalization procedures are fixed before model evaluation and released with the benchmark artifacts.
    
    \subsection{Metrics}
    
    For a fixed model configuration, we suppress the configuration index. Let $\mathcal{T}=\{1,\ldots,N\}$ denote the released state set, where $N=680$, and let $\hat{a}_t$ and $a_t^*$ denote the parsed model action and recorded reviewer action at state $t$. Recorded-Action Agreement is
    \begin{equation}
    \mathrm{RAA}
    =
    \frac{100}{N}
    \sum_{t\in\mathcal{T}}
    \mathbb{I}[\hat{a}_t=a_t^*].
    \end{equation}
    
    Define the two reference-state subsets
    \begin{equation}
    \mathcal{T}_{\mathrm{Ask}}
    =
    \{t\in\mathcal{T}:a_t^*=\mathrm{Ask}\},
    \qquad
    \mathcal{T}_{\mathrm{Proceed}}
    =
    \{t\in\mathcal{T}:a_t^*=\mathrm{Proceed}\},
    \end{equation}
    where $|\mathcal{T}_{\mathrm{Ask}}|=583$ and $|\mathcal{T}_{\mathrm{Proceed}}|=97$. The branch-specific recalls are
    \begin{align}
    \mathrm{AskR}
    &=
    \frac{100}{|\mathcal{T}_{\mathrm{Ask}}|}
    \sum_{t\in\mathcal{T}_{\mathrm{Ask}}}
    \mathbb{I}[\hat{a}_t=\mathrm{Ask}],\\
    \mathrm{ProceedR}
    &=
    \frac{100}{|\mathcal{T}_{\mathrm{Proceed}}|}
    \sum_{t\in\mathcal{T}_{\mathrm{Proceed}}}
    \mathbb{I}[\hat{a}_t=\mathrm{Proceed}].
    \end{align}
    Balanced Recorded-Action Agreement gives equal weight to the two reference branches:
    \begin{equation}
    \mathrm{BAcc}
    =
    \frac{\mathrm{AskR}+\mathrm{ProceedR}}{2}.
    \end{equation}
    
    For each $t\in\mathcal{T}_{\mathrm{Ask}}$, define the request-alignment credit
    \begin{equation}
    z_t=
    \begin{cases}
    \displaystyle
    \max_{e\in\mathcal{E}_t^+}
    \ell(\hat{q}_t,e),
    & \text{if }\hat{a}_t=\mathrm{Ask}\text{ and }\hat{q}_t\text{ is valid},\\
    0,
    & \text{otherwise},
    \end{cases}
    \end{equation}
    where
    \begin{equation}
    \ell(\hat{q}_t,e)=
    \begin{cases}
    1, & \text{direct alignment with the evidence need},\\
    0.5, & \text{relevant but incomplete alignment},\\
    0, & \text{unrelated request}.
    \end{cases}
    \end{equation}
    
    Evidence-Request Alignment measures end-to-end request alignment over all recorded Ask states:
    \begin{equation}
    \mathrm{ERA}
    =
    \frac{100}{|\mathcal{T}_{\mathrm{Ask}}|}
    \sum_{t\in\mathcal{T}_{\mathrm{Ask}}}z_t.
    \end{equation}
    
    Let
    \begin{equation}
    \mathcal{A}
    =
    \{t\in\mathcal{T}_{\mathrm{Ask}}:\hat{a}_t=\mathrm{Ask}\}
    \end{equation}
    denote the recorded Ask states on which the model enters the Ask branch. For $|\mathcal{A}|>0$, Conditional Request Alignment is
    \begin{equation}
    \mathrm{CRA}
    =
    \frac{100}{|\mathcal{A}|}
    \sum_{t\in\mathcal{A}}z_t.
    \end{equation}
    When $|\mathcal{A}|=0$, CRA is undefined and is reported as ``--''. For $|\mathcal{A}|>0$, the shared recorded-Ask denominator yields the exact identity
    \begin{equation}
    \mathrm{ERA}
    =
    \frac{\mathrm{AskR}\times\mathrm{CRA}}{100}.
    \end{equation}
    
    Direct Evidence Hit and Relevant Evidence Hit are
    \begin{align}
    \mathrm{DEH}
    &=
    \frac{100}{|\mathcal{T}_{\mathrm{Ask}}|}
    \sum_{t\in\mathcal{T}_{\mathrm{Ask}}}
    \mathbb{I}[z_t=1],\\
    \mathrm{REH}
    &=
    \frac{100}{|\mathcal{T}_{\mathrm{Ask}}|}
    \sum_{t\in\mathcal{T}_{\mathrm{Ask}}}
    \mathbb{I}[z_t\geq0.5].
    \end{align}
    Because $z_t\in\{0,0.5,1\}$ under the three-level rubric,
    \begin{equation}
    \mathrm{ERA}
    =
    \frac{\mathrm{DEH}+\mathrm{REH}}{2}.
    \end{equation}
    
    \subsection{Interpretation Boundary}
    
    FinRisk-Ask evaluates recorded professional transitions and trajectory-supported evidence needs under an offline replay setting. Later observations establish target provenance, but the evaluation does not execute generated requests or estimate their causal value after acquisition.
    
    The target set represents evidence needs recoverable from completed trajectories and does not enumerate every professionally reasonable acquisition strategy. Similarly, the recorded Ask-or-Proceed action reflects one operational workflow and should not be interpreted as the unique optimal review policy.
    
    Therefore, BAcc measures agreement with recorded workflow behavior, while ERA and CRA measure alignment with the released evidence-target set. These metrics characterize evidence-state control under the benchmark protocol rather than provide a complete measure of professional review quality.

\section{Supplementary Capability Analyses}\label{app:supplementary-analysis}

    This appendix provides additional analyses supporting the main empirical claims in Section~\ref{sec:experiments}: operation-specific model profiles, the limitations of knowledge-based routing, and the separation between recorded action agreement and evidence-request targeting.
    
    \subsection{Robustness of Operation-Specific Capability Profiles}
    
    The main experiments show that downstream review operations induce different model rankings. We further examine whether this observation is driven only by large differences in broad Domain Knowledge performance or whether models with similar knowledge scores can still exhibit different operational profiles.
    
    For each configuration pair $(m,n)$, we define the absolute difference in Domain Knowledge macro performance as
    \begin{equation}
    D_{mn}^{K}=|K_m-K_n|.
    \end{equation}
    
    Let $M=|\mathcal{M}_{\mathrm{cmp}}|=32$. For each downstream operation $o$, let $R_{m,o}$ denote the descending average rank of configuration $m$ within $\mathcal{M}_{\mathrm{cmp}}$, with rank 1 assigned to the highest score and tied configurations receiving average ranks. We map this rank to a within-operation percentile:
    \begin{equation}
    P_{m,o}=100\frac{M-R_{m,o}}{M-1}.
    \end{equation}
    
    The downstream profile gap between configurations $m$ and $n$ is
    \begin{equation}
    G_{mn}=\frac{1}{|\mathcal{O}|}\sum_{o\in\mathcal{O}}|P_{m,o}-P_{n,o}|,
    \end{equation}
    where $\mathcal{O}$ denotes the eleven downstream operations. Percentile normalization is used only to compare relative profiles across operations with heterogeneous native scoring scales.
    
    For the spectral analysis, let $\mathbf{C}\in\mathbb{R}^{11\times11}$ denote the Spearman rank-correlation matrix of the downstream operation scores, with eigenvalues $\lambda_1\geq\lambda_2\geq\cdots\geq\lambda_{11}\geq0$. The contribution of component $j$ is measured by $\lambda_j/\sum_{\ell=1}^{11}\lambda_\ell$.
    
    \begin{figure*}[t]
    \centering
    \includegraphics[width=0.98\textwidth]{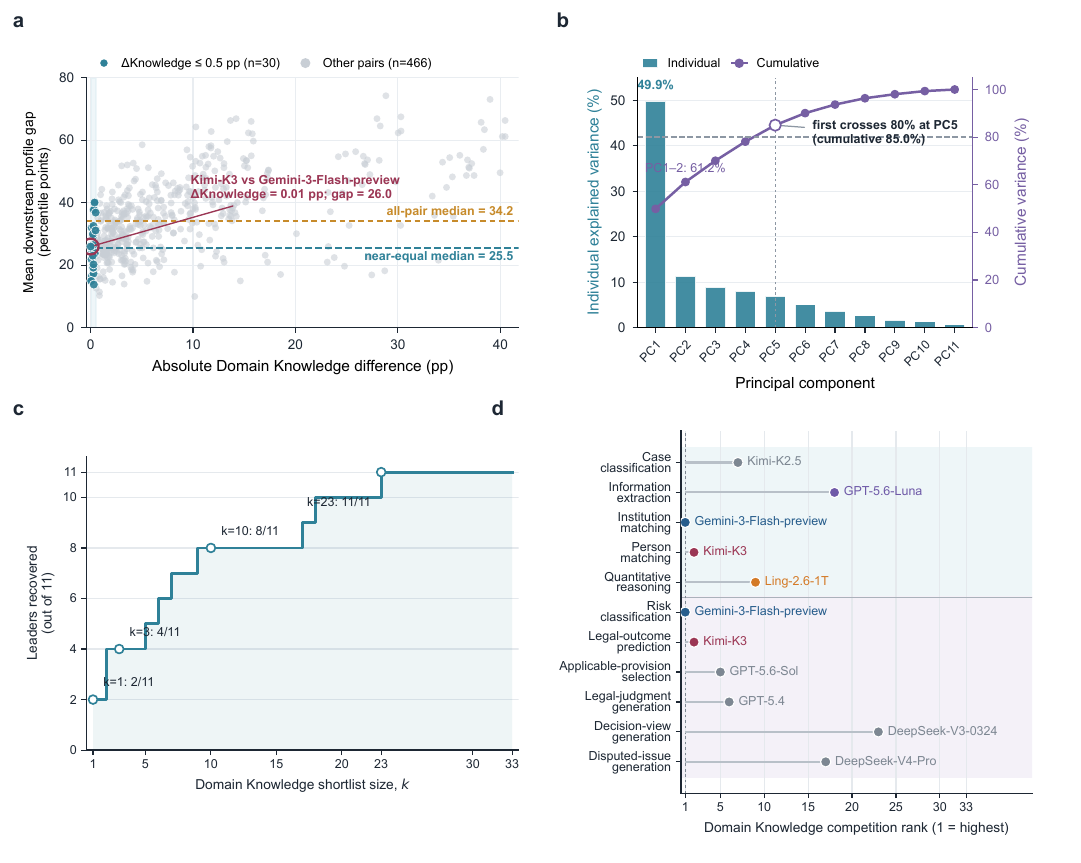}
    \caption{\textbf{Additional analyses of operation-specific capability profiles.} (a) Relationship between Domain Knowledge difference and downstream profile gap over 496 configuration pairs. Knowledge-near pairs ($|\Delta K|\leq 0.5$) still exhibit substantial downstream divergence. (b) Eigenspectrum of the eleven-operation Spearman correlation matrix, showing shared but non-identical ranking structure across operations. (c) Number of observed operation leaders recovered by Domain Knowledge-based shortlists of different sizes. (d) Domain Knowledge rank of each configuration attaining an observed operation maximum. All maxima correspond to point estimates from the evaluated configuration pool.}
    \label{fig:supplementary-capability-analysis}
    \end{figure*}
    
    Among the 30 configuration pairs with $D_{mn}^{K}\leq0.5$, the median downstream profile gap is 25.5 percentile points, demonstrating that similar broad knowledge performance does not imply similar operational behavior. Across all 496 pairs, the median profile gap is 34.2 percentile points. The eigenspectrum further shows that operation rankings share common structure but are not reducible to a single latent ordering: the leading eigenvalue accounts for 49.9\% of the total eigenvalue mass, while the first five eigenvalues account for 85.0\% cumulatively.
    
    We additionally examine whether Domain Knowledge ranking can recover operation-specific leaders. Within their respective operation-eligible configuration sets, the observed leaders occupy Domain Knowledge ranks from 1 to 23. A shortlist of size one recovers 2 of the 11 operation maxima, while a shortlist of size ten recovers 8 of 11. Full recovery requires a shortlist of size 23. These results show that broad knowledge ranking provides useful but incomplete information for operation-specific model selection.
    
    \subsection{Relationship Between Static Capability and Evidence-State Control}
    
    FinRisk-Ask evaluates evidence-state control separately from fixed-evidence operation execution. We further analyze whether static downstream capabilities are strongly associated with active-review behavior.
    
\begin{figure*}[t]
    \centering
    \includegraphics[width=0.98\textwidth]{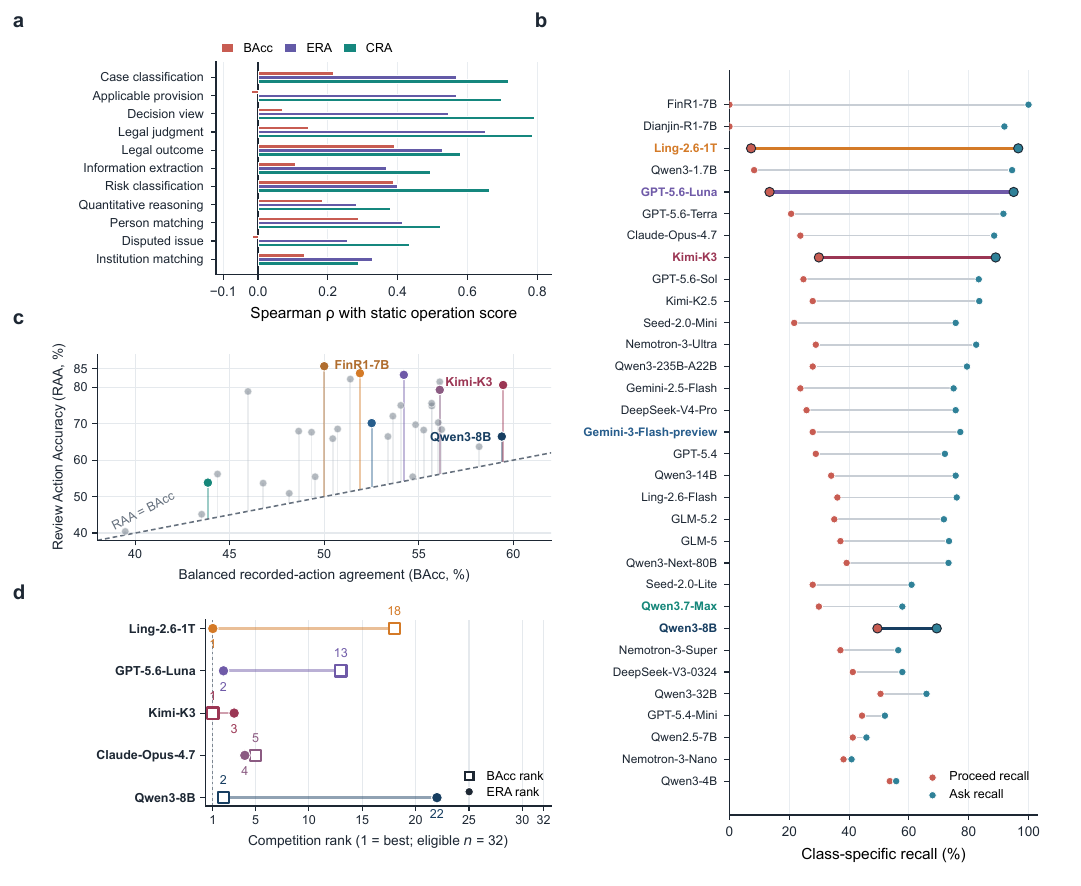}
    \caption{\textbf{Additional analyses of static capability and evidence-state behavior.} (a) Spearman associations between downstream operation scores and FinRisk-Ask metrics. (b) Ask recall and Proceed recall across non-evaluator configurations, showing asymmetric reproduction of recorded transitions. (c) Relationship between Recorded-Action Agreement (RAA) and Balanced Recorded-Action Agreement (BAcc). (d) Rank displacement between BAcc and Evidence-Request Alignment (ERA) for representative configurations.}
    \label{fig:supplementary-active-review}
    \end{figure*}

    Figure~\ref{fig:supplementary-active-review}(a) reports Spearman associations between each downstream static operation and three FinRisk-Ask metrics: Balanced Recorded-Action Agreement (BAcc), Evidence-Request Alignment (ERA), and Conditional Request Alignment (CRA). These correlations are descriptive associations within the evaluated configuration pool and do not imply causal relationships.
    
    Several applied-review operations exhibit stronger association with request targeting than with recorded-action agreement. For example, legal-judgment generation shows stronger correlation with CRA ($\rho=0.78$) than with BAcc ($\rho=0.14$), while decision-view generation shows stronger correlation with CRA ($\rho=0.79$) than with BAcc ($\rho=0.07$). These observations suggest that generating useful evidence requests relies on capabilities that are not fully captured by matching the recorded Ask-or-Proceed boundary.
    
    Figure~\ref{fig:supplementary-active-review}(b--d) further examines action-level behavior. Across the 32 non-evaluator configurations, Ask recall is generally higher than Proceed recall, indicating an asymmetric tendency toward reproducing evidence requests rather than recorded advancement decisions. Figure~\ref{fig:supplementary-active-review}(c) shows that overall RAA can differ from balanced action agreement because the two reference classes are imbalanced. Figure~\ref{fig:supplementary-active-review}(d) illustrates that configurations can shift substantially in ranking depending on whether they are evaluated by action agreement or end-to-end request alignment.
    
    Together, these analyses reinforce that FinRisk-Ask measures multiple aspects of evidence-state control: reproducing a recorded transition, recognizing when additional evidence is required, and requesting evidence aligned with the unresolved decision need.

\section{Complete Model Results}
\label{app:full-results}

    This appendix reports the complete configuration-level results underlying the analyses in Section~\ref{sec:experiments}. The tables provide the full score matrices for all benchmark components under the same evaluation protocols used in the main experiments. For semantically evaluated tasks, the evaluator configuration's own score is retained for transparency but excluded from best-value annotations and comparative analyses restricted to $\mathcal{M}_{\mathrm{cmp}}$.

    \subsection{Complete Domain Knowledge Results}
    
    The Domain Knowledge score is the unweighted macro-average over 42 knowledge task families. Each family contributes equally, preventing taxonomy groups with larger numbers of instances from dominating the overall knowledge score.
    
    \begin{table*}[t]
        \centering
        \scriptsize
        \setlength{\tabcolsep}{2.4pt}
        \caption{Complete Domain Knowledge macro results over all 33 archived configurations (\%).}
        \label{tab:full-knowledge-results}
        \begin{tabular}{lrrrrrr}
        \toprule
        \textbf{Model}&
        \textbf{Risk \& Compliance}&
        \textbf{Trade}&
        \textbf{FinTech \& Security}&
        \textbf{Commerce \& Payment}&
        \textbf{Finance \& Law}&
        \textbf{Overall}\\
        \midrule
        Dianjin-R1-7B&64.00&63.43&70.25&72.50&68.00&65.00\\
        FinR1-7B&66.35&68.00&78.25&75.50&76.00&68.93\\
        Ling-2.6-Flash&71.80&74.71&80.00&79.00&77.50&74.17\\
        Ling-2.6-1T&75.22&78.96&83.22&78.82&80.49&77.65\\
        Seed-2.0-Mini&73.22&76.64&80.43&78.32&80.04&75.62\\
        Seed-2.0-Lite&73.70&80.36&78.78&81.51&78.55&77.01\\
        Nemotron-3-Nano&49.06&49.30&57.49&57.46&49.92&50.38\\
        Nemotron-3-Super&64.74&68.49&73.37&70.36&72.16&67.43\\
        Nemotron-3-Ultra&71.81&76.89&78.06&79.73&81.85&74.96\\
        Qwen2.5-7B&62.23&61.79&67.11&71.85&70.94&63.42\\
        Qwen3-1.7B&40.13&37.87&43.78&44.49&46.05&40.21\\
        Qwen3-4B&63.17&60.38&68.45&69.46&69.10&63.32\\
        Qwen3-8B&67.12&65.68&71.24&74.42&73.06&67.66\\
        Qwen3-14B&67.92&70.93&76.49&73.66&77.65&70.48\\
        Qwen3-32B&70.90&74.46&79.73&79.20&80.66&73.79\\
        Qwen3-Next-80B&74.20&78.97&82.52&79.03&82.08&77.19\\
        Qwen3-235B-A22B&73.25&79.02&83.02&82.08&81.26&76.90\\
        Qwen3.7-Max&77.15&82.49&85.04&86.01&83.65&80.41\\
        DeepSeek-V3-0324&70.44&75.65&78.33&74.99&79.48&73.57\\
        DeepSeek-V4-Flash&71.33&75.49&79.98&76.30&77.65&74.08\\
        DeepSeek-V4-Pro&72.11&76.87&82.73&81.55&79.88&75.53\\
        Kimi-K2.5&75.23&81.95&82.15&80.25&79.21&78.56\\
        Kimi-K3&\textbf{78.11}&83.05&84.07&82.80&80.95&80.68\\
        GLM-5&73.53&78.14&80.27&79.93&78.81&76.26\\
        GLM-5.2&74.65&79.18&80.33&79.43&79.02&77.13\\
        GPT-5.4-Mini&74.38&79.48&80.59&84.34&80.16&77.42\\
        GPT-5.4&74.46&81.17&84.69&84.34&\textbf{85.34}&78.66\\
        GPT-5.6-Luna&72.49&77.61&77.59&78.81&78.53&75.27\\
        GPT-5.6-Terra&74.87&81.58&81.19&82.73&80.24&78.34\\
        GPT-5.6-Sol&76.30&83.00&80.23&82.42&76.71&79.21\\
        Gemini-2.5-Flash&70.27&77.68&80.56&80.51&77.98&74.57\\
        Gemini-3-Flash-preview&77.24&\textbf{83.80}&84.33&86.41&80.53&\textbf{80.69}\\
        Claude-Opus-4.7&76.55&82.15&\textbf{86.92}&\textbf{86.81}&82.75&80.19\\
        \bottomrule
        \end{tabular}
    \end{table*}
    
    \subsection{Complete Fixed-Evidence Operation Results}
    
    Table~\ref{tab:full-operation-results} reports the complete results for Evidence-Grounded Processing and structured-output Applied Review operations. Columns correspond to the downstream operations defined in Table~\ref{tab:workflow-map}. Each operation retains its native scoring protocol and is not combined into a single downstream aggregate score.

    \begin{table*}[t]
        \centering
        \scriptsize
        \setlength{\tabcolsep}{1.8pt}
        \caption{Complete Evidence-Grounded Processing and structured Applied Review results over all 33 archived configurations (\%). Column abbreviations correspond to the operations summarized in Table~\ref{tab:workflow-map}; their full evaluation contracts are reported in Table~\ref{tab:operation-contracts}.}
        \label{tab:full-operation-results}
        \begin{tabular*}{\textwidth}{@{\extracolsep{\fill}}lrrrrrrrr@{}}
        \toprule
        \textbf{Model}&
        \textbf{Case}&
        \textbf{Extract.}&
        \textbf{Inst.}&
        \textbf{Person}&
        \textbf{Quant.}&
        \textbf{Risk}&
        \textbf{Legal}&
        \textbf{Provision}\\
        \midrule
        Dianjin-R1-7B&77.00&59.00&90.00&70.00&63.00&31.00&38.00&33.00\\
        FinR1-7B&76.00&10.00&82.00&82.00&73.00&29.00&64.00&35.00\\
        Ling-2.6-Flash&80.00&72.00&78.00&80.00&62.00&75.00&60.00&40.00\\
        Ling-2.6-1T&82.92&74.53&78.80&80.40&\textbf{86.21}&75.68&64.76&52.10\\
        Seed-2.0-Mini&83.15&75.42&86.80&73.80&79.31&66.61&62.22&36.95\\
        Seed-2.0-Lite&85.62&84.48&84.00&81.80&79.31&73.22&63.81&56.40\\
        Nemotron-3-Nano&66.29&70.43&75.60&73.40&65.52&54.92&53.76&23.50\\
        Nemotron-3-Super&79.78&72.44&82.40&79.20&75.00&71.86&56.72&30.15\\
        Nemotron-3-Ultra&82.02&81.49&74.80&85.60&81.90&75.41&61.69&29.45\\
        Qwen2.5-7B&76.63&59.14&78.00&82.00&75.00&53.28&62.96&40.55\\
        Qwen3-1.7B&54.61&61.79&79.60&76.40&56.03&71.86&58.94&16.20\\
        Qwen3-4B&76.40&75.72&80.00&84.00&75.00&71.31&62.33&12.80\\
        Qwen3-8B&81.75&61.69&80.32&83.00&70.86&74.59&64.49&17.44\\
        Qwen3-14B&81.80&76.82&78.80&85.40&74.14&68.85&64.97&33.40\\
        Qwen3-32B&82.11&74.63&90.00&84.36&79.66&73.77&62.96&27.44\\
        Qwen3-Next-80B&84.94&76.02&90.80&81.60&80.17&73.77&65.71&41.30\\
        Qwen3-235B-A22B&83.37&81.25&88.00&83.40&81.03&62.84&65.08&40.50\\
        Qwen3.7-Max&85.39&82.79&84.00&84.40&78.45&75.68&66.03&46.85\\
        DeepSeek-V3-0324&83.15&79.12&84.40&80.40&76.72&74.86&63.70&34.15\\
        DeepSeek-V4-Flash&82.92&51.84&88.00&83.40&72.41&75.41&63.70&53.00\\
        DeepSeek-V4-Pro&85.62&76.02&86.80&85.20&68.97&70.49&64.97&41.45\\
        Kimi-K2.5&\textbf{88.09}&84.28&87.20&83.60&75.86&56.56&65.93&46.40\\
        Kimi-K3&87.19&74.43&84.80&\textbf{87.00}&77.59&74.86&\textbf{68.15}&51.65\\
        GLM-5&84.72&81.39&87.60&84.20&84.48&73.22&64.76&51.50\\
        GLM-5.2&84.94&81.69&85.60&83.60&78.45&71.86&65.71&51.15\\
        GPT-5.4-Mini&85.62&83.78&81.60&85.80&76.72&71.86&64.97&55.35\\
        GPT-5.4&86.29&83.98&82.00&85.60&75.86&74.04&67.51&60.45\\
        GPT-5.6-Luna&84.04&\textbf{85.97}&85.60&85.80&75.00&71.86&65.93&50.80\\
        GPT-5.6-Terra&85.62&83.48&90.40&85.80&76.72&73.77&67.83&56.95\\
        GPT-5.6-Sol&87.42&85.07&90.00&86.00&77.59&74.59&67.09&\textbf{67.55}\\
        Gemini-2.5-Flash&82.25&75.72&89.20&85.60&64.66&72.95&60.11&66.25\\
        Gemini-3-Flash-preview&84.94&67.96&\textbf{93.60}&85.80&75.00&\textbf{76.78}&59.68&65.75\\
        Claude-Opus-4.7&85.17&76.62&63.20&50.80&73.28&75.14&65.19&54.80\\
        \bottomrule
        \end{tabular*}
    \end{table*}
    
    \subsection{Complete Open-Ended Applied Review Results}
    Table~\ref{tab:full-generation-results} reports the complete results for open-ended Applied Review operations evaluated with the fixed semantic evaluator described in Appendix~\ref{app:protocol}. The evaluator configuration's self-score is retained for transparency and marked with $\dagger$, but excluded from comparative analyses involving judge-scored operations.

    \begin{table*}[t]
        \centering
        \scriptsize
        \setlength{\tabcolsep}{4.5pt}
        \caption{Complete open-ended Applied Review results over all 33 archived configurations (\%). DeepSeek-V4-Flash is the fixed semantic evaluator. Its self-scored row is marked with $\dagger$ and excluded from best-value annotation and evaluator-dependent comparisons.}
        \label{tab:full-generation-results}
        \begin{tabular}{lrrr}
        \toprule
        \textbf{Model}&
        \textbf{Legal Judgment}&
        \textbf{Decision View}&
        \textbf{Disputed Issue}\\
        \midrule
        Dianjin-R1-7B&29.00&20.00&36.00\\
        FinR1-7B&18.00&15.00&52.00\\
        Ling-2.6-Flash&30.00&37.00&63.00\\
        Ling-2.6-1T&40.27&55.69&63.30\\
        Seed-2.0-Mini&28.95&47.11&72.73\\
        Seed-2.0-Lite&29.66&58.73&73.35\\
        Nemotron-3-Nano&21.48&17.09&39.57\\
        Nemotron-3-Super&28.62&37.17&76.67\\
        Nemotron-3-Ultra&30.68&52.00&71.58\\
        Qwen2.5-7B&28.10&36.24&70.40\\
        Qwen3-1.7B&16.18&7.76&41.13\\
        Qwen3-4B&19.38&21.99&68.83\\
        Qwen3-8B&17.37&30.05&37.30\\
        Qwen3-14B&35.02&39.03&53.81\\
        Qwen3-32B&34.54&35.60&53.30\\
        Qwen3-Next-80B&32.43&53.10&67.51\\
        Qwen3-235B-A22B&39.26&49.00&57.35\\
        Qwen3.7-Max&36.22&57.84&70.58\\
        DeepSeek-V3-0324&36.26&\textbf{71.31}&51.68\\
        DeepSeek-V4-Flash$^\dagger$&25.16&54.86&81.23\\
        DeepSeek-V4-Pro&30.67&54.48&\textbf{77.12}\\
        Kimi-K2.5&34.49&62.74&61.26\\
        Kimi-K3&35.05&64.13&74.84\\
        GLM-5&31.60&57.16&72.49\\
        GLM-5.2&31.59&58.59&68.37\\
        GPT-5.4-Mini&39.86&48.77&62.13\\
        GPT-5.4&\textbf{46.36}&62.91&54.04\\
        GPT-5.6-Luna&41.70&52.77&61.33\\
        GPT-5.6-Terra&40.98&54.72&58.29\\
        GPT-5.6-Sol&42.09&62.25&58.11\\
        Gemini-2.5-Flash&45.29&65.27&74.05\\
        Gemini-3-Flash-preview&40.48&64.61&74.69\\
        Claude-Opus-4.7&35.53&51.97&74.06\\
        \bottomrule
        \end{tabular}
    \end{table*}
    
    \subsection{Complete FinRisk-Ask Results}
    
    Table~\ref{tab:full-finriskask-results} reports complete FinRisk-Ask results over all evaluated configurations. The metrics follow the definitions in Appendix~\ref{app:finriskask-protocol}: RAA measures agreement with recorded actions, BAcc balances Ask and Proceed recall, ERA measures end-to-end evidence-request alignment, and CRA measures request alignment conditional on entering the Ask branch. The request-alignment evaluator configuration is marked with $\dagger$ and excluded from evaluator-dependent comparative analyses.
    
\begin{table*}[t]
    \centering
    \scriptsize
    \setlength{\tabcolsep}{2.8pt}
    \renewcommand{\arraystretch}{1.06}
    \caption{\textbf{Complete FinRisk-Ask results for two reference policies and 33 archived model configurations (\%).} ERA is the end-to-end evidence-acquisition metric. DeepSeek-V4-Flash is the fixed request-alignment evaluator; its self-scored row is marked with $\dagger$ and excluded from evaluator-dependent comparisons.}
    \label{tab:full-finriskask-results}
    \begin{tabular}{l>{\columncolor{YaleBlue!5}}r@{\hspace{3pt}}rrrr@{\hspace{3pt}}rrr}
    \toprule
    & \multicolumn{1}{>{\columncolor{YaleBlue!5}}c}{\textbf{End-to-end}}
    & \multicolumn{4}{c}{\textbf{Action behavior}}
    & \multicolumn{3}{c}{\textbf{Request alignment}}\\
    \cmidrule(lr){2-2}\cmidrule(lr){3-6}\cmidrule(l){7-9}
    \textbf{Model} & \textbf{ERA} & \textbf{BAcc} & \textbf{AskR} & \textbf{ProceedR} & \textbf{RAA} & \textbf{CRA} & \textbf{DEH} & \textbf{REH}\\
    \midrule
    \multicolumn{1}{@{}l}{\textit{Reference policies}} & & & & & & & & \\
    Always-Ask&--&50.00&100.00&0.00&85.74&--&--&--\\
    Always-Proceed&--&50.00&0.00&100.00&14.26&--&--&--\\
    \midrule
    \multicolumn{1}{@{}l}{\textit{Model configurations}} & & & & & & & & \\
    Dianjin-R1-7B&57.72&45.97&91.94&0.00&78.82&62.78&51.97&63.46\\
    FinR1-7B&48.46&50.00&\textbf{100.00}&0.00&\textbf{85.74}&48.46&45.45&51.46\\
    Ling-2.6-Flash&51.80&56.03&75.99&36.08&70.29&68.17&49.39&54.20\\
    Ling-2.6-1T&\textbf{79.75}&51.89&96.57&7.22&83.82&82.58&\textbf{76.50}&\textbf{83.01}\\
    Seed-2.0-Mini&54.54&48.65&75.64&21.65&67.94&72.10&52.31&56.77\\
    Seed-2.0-Lite&42.96&44.36&60.89&27.84&56.18&70.55&40.30&45.62\\
    Nemotron-3-Nano&24.78&39.48&40.82&38.14&40.44&60.70&22.98&26.58\\
    Nemotron-3-Super&36.27&46.77&56.43&37.11&53.68&64.27&34.99&37.56\\
    Nemotron-3-Ultra&67.75&55.69&82.50&28.87&74.85&82.12&66.20&69.29\\
    Qwen2.5-7B&26.84&43.52&45.80&41.24&45.15&58.61&25.04&28.64\\
    Qwen3-1.7B&31.38&51.38&94.51&8.25&82.21&33.20&29.33&33.44\\
    Qwen3-4B&34.30&54.68&55.75&\textbf{53.61}&55.44&61.53&32.93&35.67\\
    Qwen3-8B&48.79&59.39&69.30&49.48&66.47&70.41&46.14&51.45\\
    Qwen3-14B&50.77&54.83&75.64&34.02&69.71&67.12&48.88&52.65\\
    Qwen3-32B&48.54&58.19&65.87&50.52&63.68&73.69&45.96&51.11\\
    Qwen3-Next-80B&57.71&56.21&73.24&39.18&68.38&78.79&56.08&59.34\\
    Qwen3-235B-A22B&62.86&53.63&79.42&27.84&72.06&79.15&61.57&64.15\\
    Qwen3.7-Max&50.68&43.85&57.80&29.90&53.82&\textbf{87.67}&49.91&51.45\\
    DeepSeek-V3-0324&45.79&49.52&57.80&41.24&55.44&79.22&44.42&47.16\\
    DeepSeek-V4-Flash$^\dagger$&59.51&50.19&81.82&18.56&72.79&72.73&57.63&61.40\\
    DeepSeek-V4-Pro&60.29&50.71&75.64&25.77&68.53&79.70&58.49&62.09\\
    Kimi-K2.5&66.80&55.68&83.53&27.84&75.59&79.97&65.35&68.26\\
    Kimi-K3&77.10&\textbf{59.46}&89.02&29.90&80.59&86.61&\textbf{76.50}&77.70\\
    GLM-5&55.66&55.26&73.41&37.11&68.24&75.82&54.20&57.11\\
    GLM-5.2&58.49&53.37&71.70&35.05&66.47&81.58&57.28&59.69\\
    GPT-5.4-Mini&39.27&48.15&51.97&44.33&50.88&75.56&38.25&40.30\\
    GPT-5.4&59.17&50.45&72.04&28.87&65.88&82.13&57.97&60.37\\
    GPT-5.6-Luna&77.87&54.21&95.03&13.40&83.38&81.95&75.81&79.93\\
    GPT-5.6-Terra&75.64&56.11&91.60&20.62&81.47&82.58&74.09&77.18\\
    GPT-5.6-Sol&67.75&54.05&83.36&24.74&75.00&81.27&66.55&68.95\\
    Gemini-2.5-Flash&62.60&49.33&74.96&23.71&67.65&83.51&61.23&63.97\\
    Gemini-3-Flash-preview&64.49&52.51&77.19&27.84&70.15&83.55&62.95&66.03\\
    Claude-Opus-4.7&75.98&56.11&88.51&23.71&79.26&85.85&74.09&77.87\\
    \bottomrule
    \end{tabular}
\end{table*}

\section{Qualitative Case Studies}
\label{app:qualitative-cases}

    This section provides qualitative examples illustrating how FinRiskAtlas instantiates operation-level evaluation contracts. The examples are not intended to compare model performance or serve as additional benchmark statistics. Instead, they demonstrate how different professional decisions require different visible evidence boundaries, reviewer artifacts, and scoring criteria. The static examples are organized according to the three benchmark layers introduced in Section~\ref{sec:benchmark}: Domain Knowledge, Evidence-Grounded Processing, and Applied Review.
    
    Each static example contains the evaluation input, the reference decision or artifact, and an example model output. The displayed outputs are observable generations returned by the model and do not represent reconstructed hidden reasoning. For open-ended Applied Review tasks, evaluation scores are produced by the fixed semantic evaluator under the corresponding rubric; the examples illustrate the evaluation contract rather than standalone evidence of general model capability.
    
    \subsection{Domain Knowledge}
    \label{app:cases-domain-knowledge}
    
\begin{figure}[!htbp]
        \centering
        \includegraphics[width=\linewidth]
        {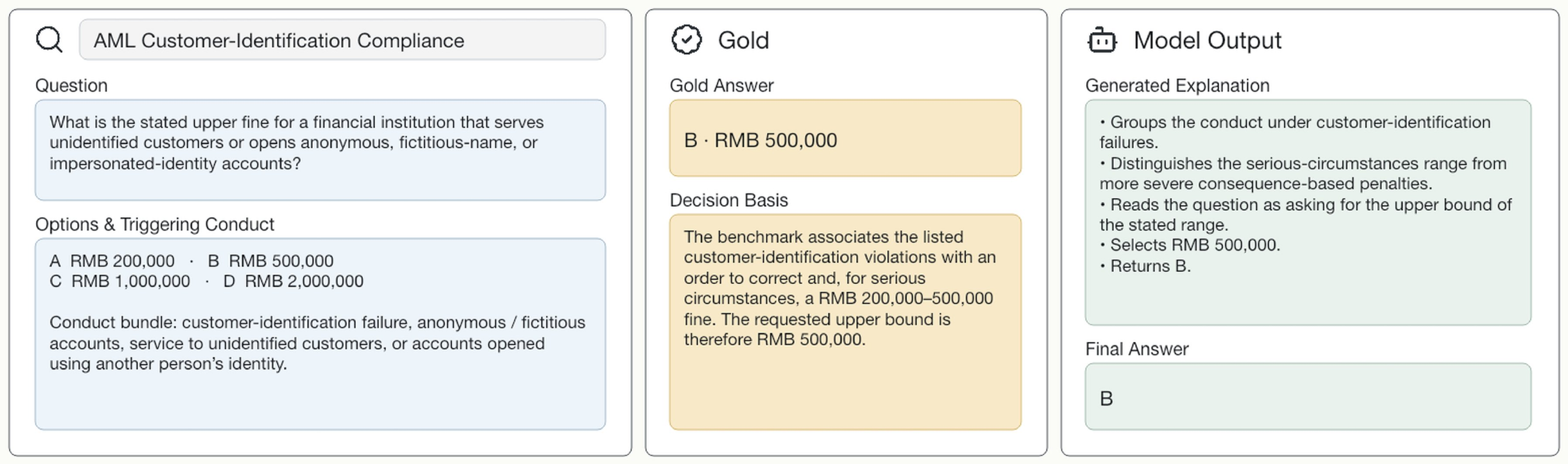}
        \caption[
            Domain Knowledge example: AML customer-identification compliance
        ]{
            \textbf{Domain Knowledge example: AML customer-identification compliance.}
            The task evaluates whether the model can retrieve the applicable customer-identification requirement and determine the stated upper bound of the corresponding administrative fine. Solving the item requires distinguishing the directly applicable penalty range from more severe sanctions associated with different consequences or aggravating conditions.
        }
        \label{fig:case-domain-aml}
    \end{figure}

    The Domain Knowledge layer measures whether models possess conceptual and regulatory knowledge required to interpret financial-review evidence. Figure~\ref{fig:case-domain-aml} presents an AML compliance example in which the model must identify the applicable customer-identification penalty range. The case illustrates that even knowledge-oriented evaluation requires a precise decision target: the model must distinguish the requested administrative fine from other possible sanctions associated with different legal consequences.
    
    \subsection{Evidence-Grounded Processing}
    \label{app:cases-evidence-processing}
    
    Evidence-Grounded Processing evaluates whether models can transform heterogeneous records into structured, decision-relevant evidence. The following examples cover four representative operations: long-context case classification, quantitative verification, structured information extraction, and entity reconciliation. Together, they illustrate that evidence processing is not a single retrieval capability but a collection of operations with different decision objects and output contracts.

    \begin{figure}[!htbp]
        \centering
        \includegraphics[width=\linewidth]
        {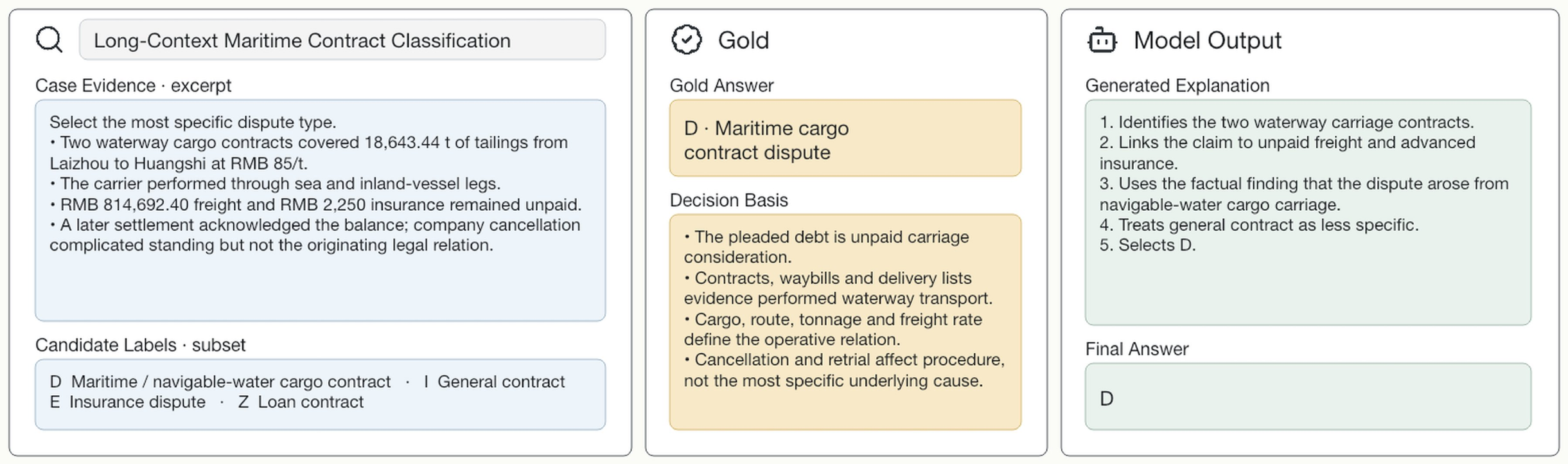}
        \caption[
            Evidence-Grounded Processing example: long-context maritime contract classification
        ]{
            \textbf{Evidence-Grounded Processing example: long-context maritime contract classification.}
            The input contains contracts, waybills, delivery records, freight and insurance amounts, performance evidence, and procedural events. The model must distinguish the operative legal relationship from contextual and procedural distractors and select the most specific case category.
        }
        \label{fig:case-evidence-maritime}
    \end{figure}
    
    \begin{figure}[!htbp]
        \centering
        \includegraphics[width=\linewidth]
        {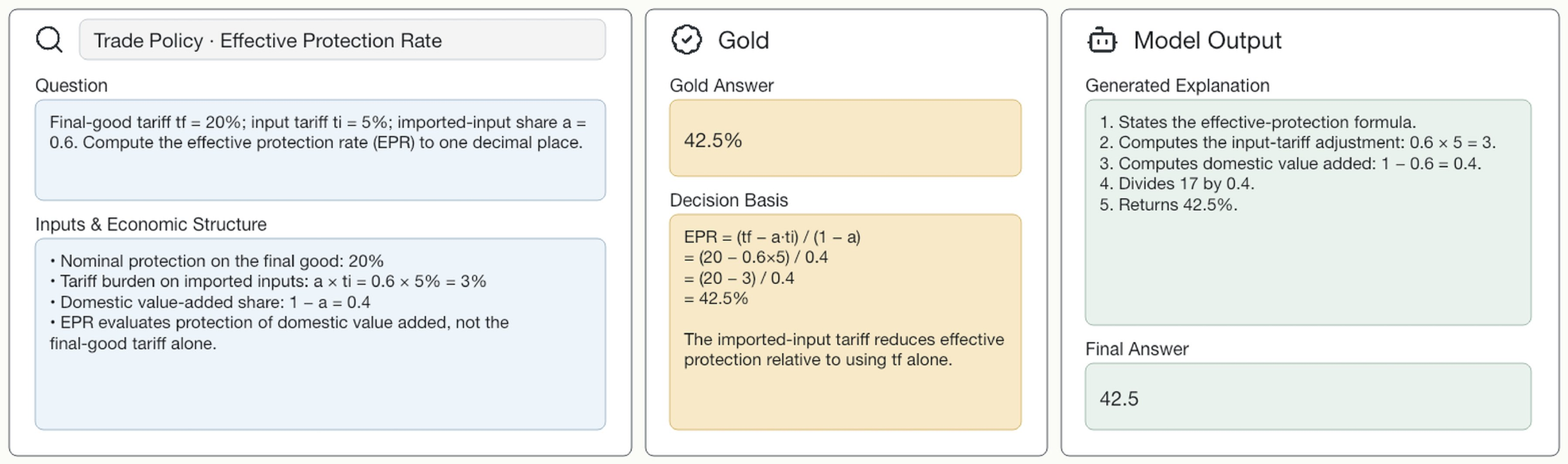}
        \caption[
            Evidence-Grounded Processing example: quantitative financial reasoning
        ]{
            \textbf{Evidence-Grounded Processing example: quantitative financial reasoning.}
            The effective-protection-rate task requires the model to identify the relevant variables, apply the correct formula, compute the tariff burden on imported inputs, and normalize by domestic value added.
        }
        \label{fig:case-evidence-epr}
    \end{figure}
    
    \begin{figure}[!htbp]
        \centering
        \includegraphics[width=\linewidth]
        {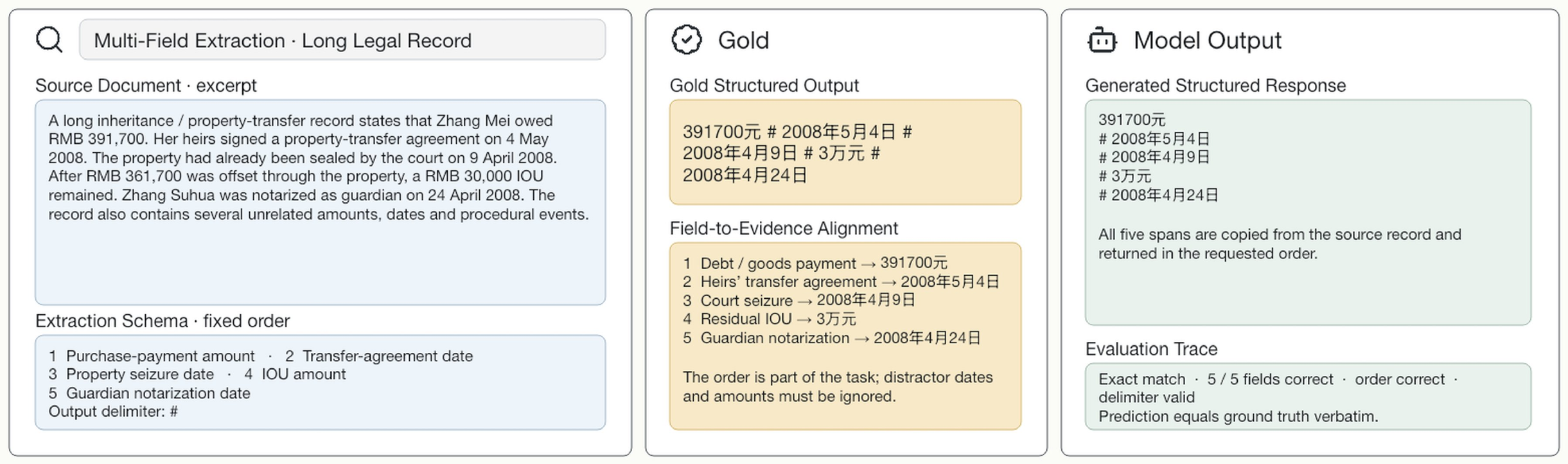}
        \caption[
            Evidence-Grounded Processing example: multi-field extraction
        ]{
            \textbf{Evidence-Grounded Processing example: multi-field extraction from a long legal record.}
            The model must extract five evidence spans in a predefined order while ignoring unrelated monetary values, dates, and procedural events. The task contract evaluates field correctness, field ordering, delimiter validity, and agreement with the reference output.
        }
        \label{fig:case-evidence-extraction}
    \end{figure}
    
    \begin{figure}[!htbp]
        \centering
        \includegraphics[width=\linewidth]
        {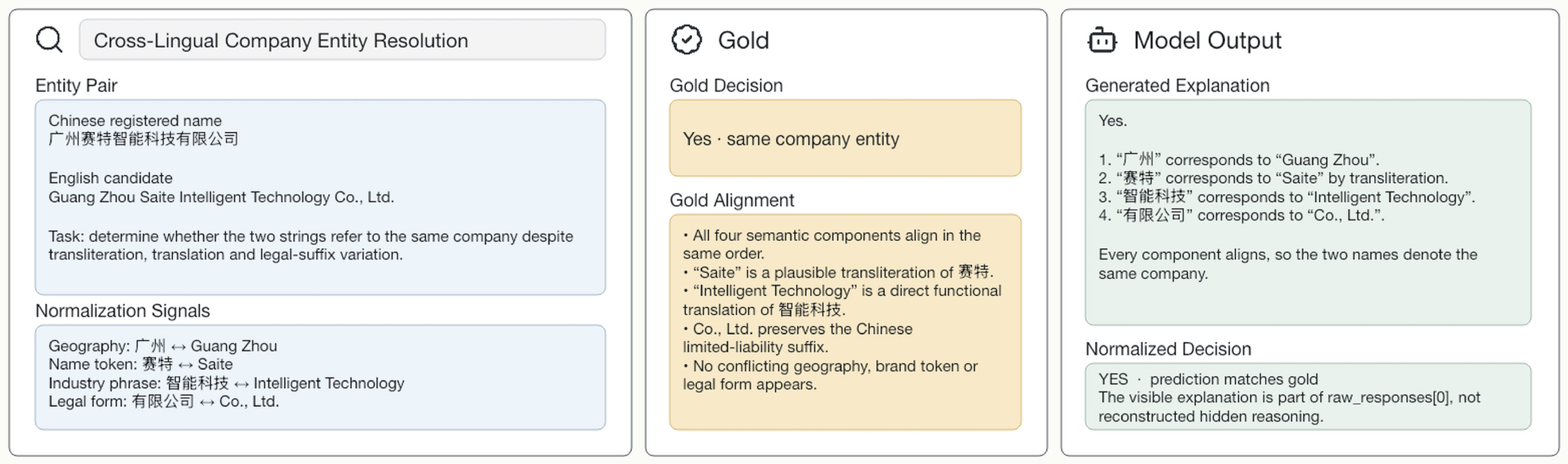}
        \caption[Evidence-Grounded Processing example: cross-lingual entity resolution]{
            \textbf{Evidence-Grounded Processing example: cross-lingual company entity resolution.}
            The task requires the model to determine whether Chinese and English company names refer to the same entity. The decision depends on the joint alignment of geographic information, transliterated name components, translated industry terms, and legal-entity suffixes.
        }
        \label{fig:case-evidence-entity}
    \end{figure}

    \subsection{Applied Review}
    \label{app:cases-applied-review}
    
    The Applied Review layer evaluates whether models can integrate evidence, apply professional decision frameworks, and generate case-level artifacts. These examples highlight a central motivation of FinRiskAtlas: the same underlying case material can support multiple professional operations, while each operation requires a different output artifact and evaluation criterion.
    
\begin{figure}[!htbp]
        \centering
        \includegraphics[width=\linewidth]
        {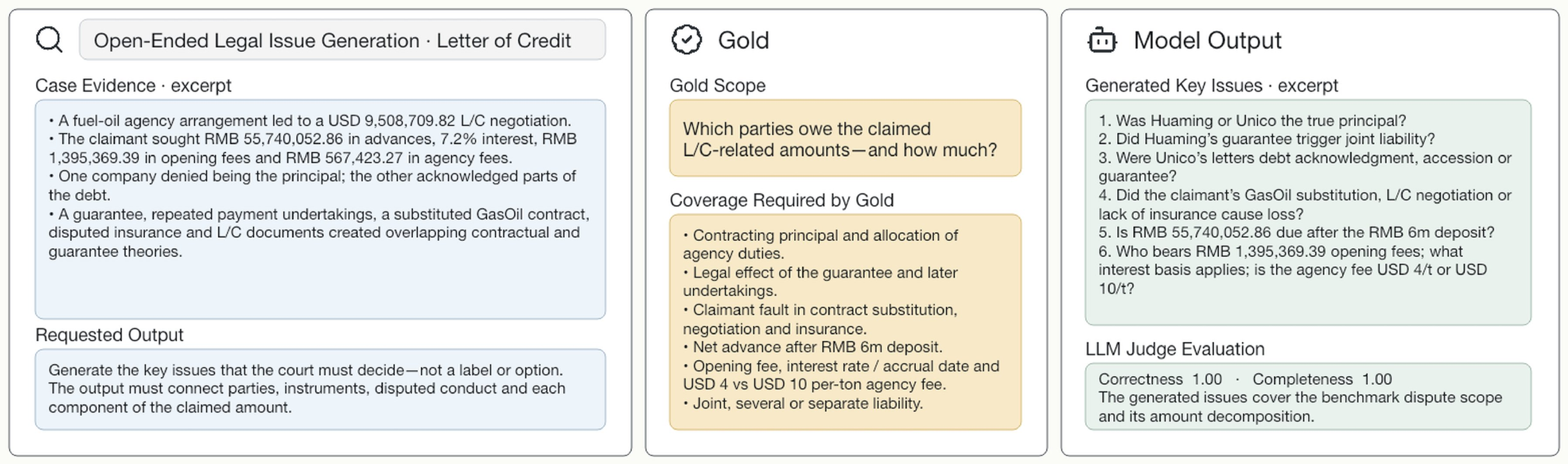}
        \caption[
            Applied Review example: open-ended disputed-issue generation
        ]{
            \textbf{Applied Review example: open-ended disputed-issue generation.}
            The source case contains overlapping agency, letter-of-credit, guarantee, insurance, and payment relationships. The model must transform these facts into a structured set of issues covering the responsible parties, contractual instruments, disputed conduct, claimed amounts, interest and fee calculations, and alternative liability relationships.
        }
        \label{fig:case-applied-issues}
    \end{figure}

\begin{figure}[!htbp]
        \centering
        \includegraphics[width=\linewidth]
        {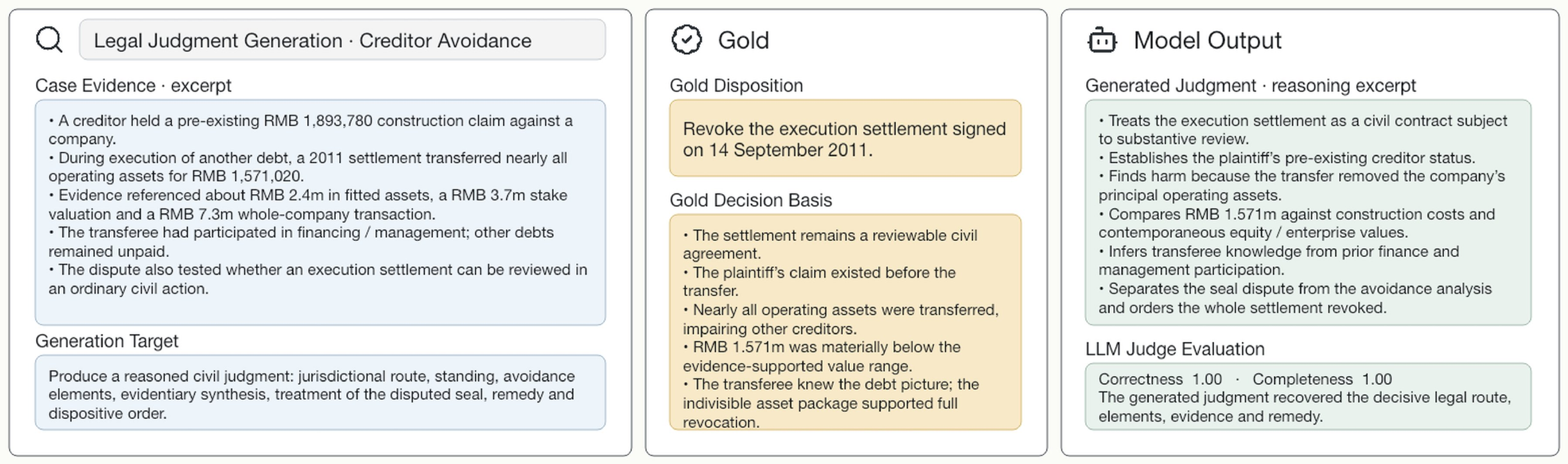}
        \caption[
            Applied Review example: legal-judgment generation
        ]{
            \textbf{Applied Review example: legal-judgment generation.}
            The model must identify the appropriate legal route, establish the claimant's standing, synthesize transaction and valuation evidence, determine whether the challenged asset transfer impaired creditors, address the disputed execution settlement, and produce an appropriate dispositive analysis.
        }
        \label{fig:case-applied-judgment}
    \end{figure}

\begin{figure}[!htbp]
        \centering
        \includegraphics[width=\linewidth]
        {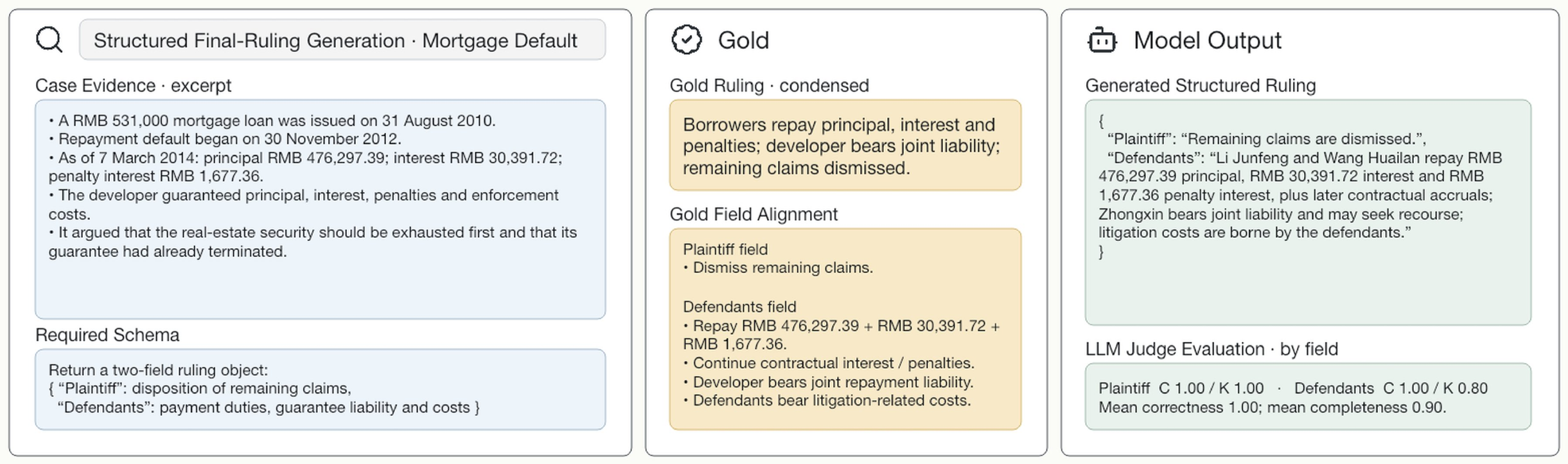}
        \caption[
            Applied Review example: structured decision-view generation
        ]{
            \textbf{Applied Review example: structured decision-view generation.}
            The model must express the supported review position in a constrained dispositive schema that preserves the affected parties, monetary obligations, and procedural treatment. This example illustrates a structured output form within the decision-view generation contract rather than an additional downstream operation.
        }
        \label{fig:case-applied-ruling}
    \end{figure}

    Figure~\ref{fig:case-applied-issues} illustrates disputed-issue generation, where the required artifact is a structured issue list rather than a final judgment. Figure~\ref{fig:case-applied-judgment} illustrates legal-judgment generation, which requires synthesizing evidence, legal elements, and remedies into a supported analysis. Figure~\ref{fig:case-applied-ruling} illustrates a structured instance of decision-view generation, where the required review position is expressed through a constrained dispositive schema rather than unrestricted prose.

    \subsection{FinRisk-Ask: Action Agreement versus Request Targeting}
    \label{app:cases-finriskask}
    
    FinRisk-Ask evaluates evidence-state control through offline replay rather than interactive dialogue. The model receives only the information available before the recorded reviewer transition. Later trajectory information is withheld during inference and is used only to construct expert-verified evaluation targets.
    
    The three examples below illustrate three request-alignment outcomes. They correspond to direct alignment, partial alignment, and missed evidence acquisition. Each example separates the visible pre-action context from the observed future evidence used only for evaluation. The examples demonstrate why entering the Ask branch and producing a useful evidence request are distinct capabilities.
    
    Where case cards display \texttt{CONTINUE} and \texttt{STOP}, \texttt{CONTINUE} denotes Ask and \texttt{STOP} denotes Proceed in the archived output protocol. These tokens represent the replayed action transition and should not be interpreted as an interactive dialogue policy or as a claim that the recorded workflow is the only valid professional strategy.

    \begin{figure}[!htbp]
        \centering
        \includegraphics[width=\linewidth]
        {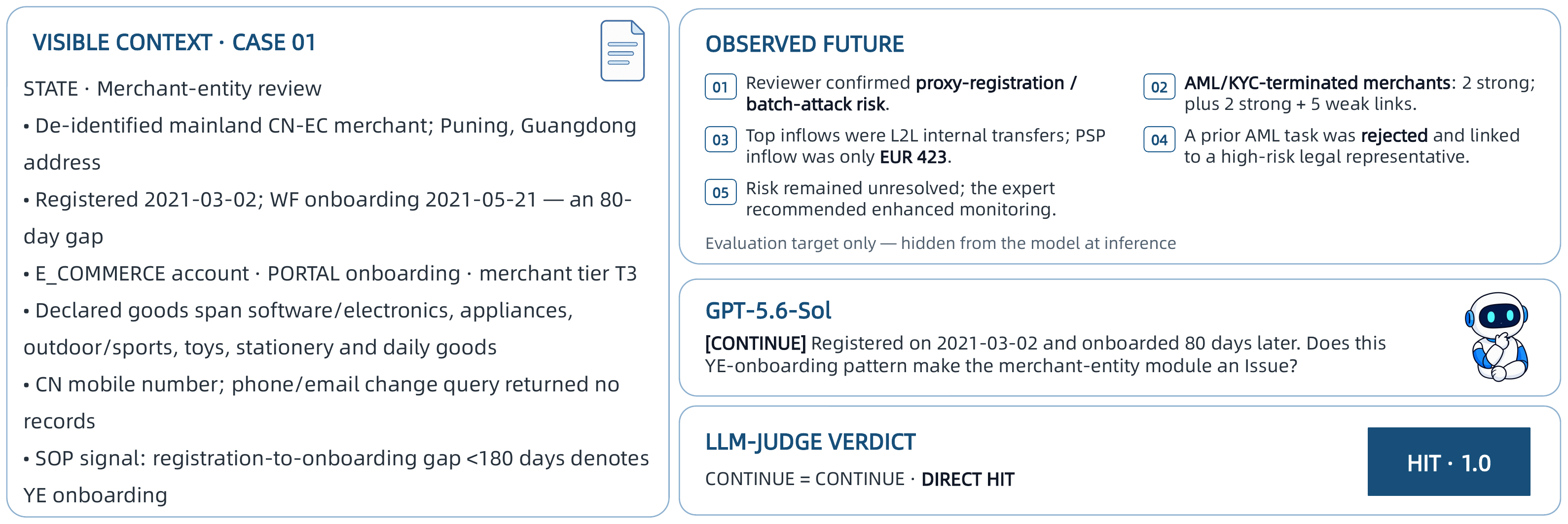}
        \caption{\textbf{Direct request alignment ($z_t=1$).}
        The visible record leaves a proxy-registration or batch-onboarding concern unresolved. The model follows the recorded Ask branch and requests verification of that concern. The request matches a retained need supported by the later trajectory, so the state is both an Ask-action hit and a direct alignment hit. The observed-future panel was unavailable at inference.}
        \label{fig:case-finriskask-direct}
    \end{figure}
    
    \begin{figure}[!htbp]
        \centering
        \includegraphics[width=\linewidth]
        {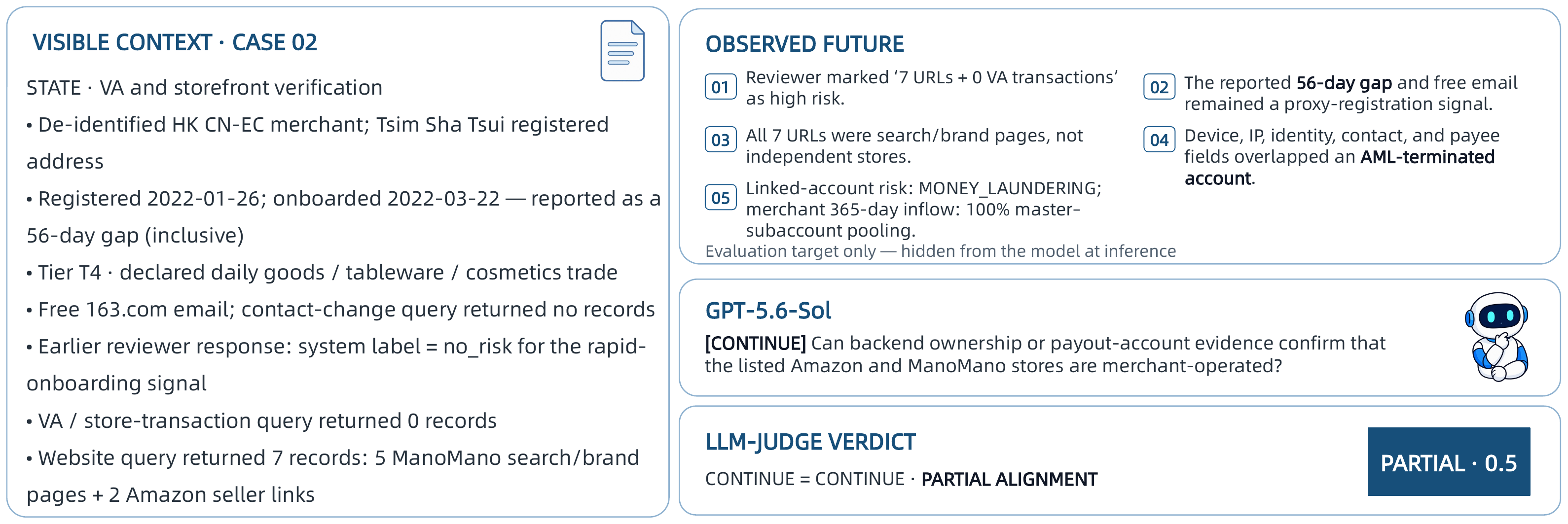}
        \caption{\textbf{Partial request alignment ($z_t=0.5$).}
        The model follows the recorded Ask branch, but the question captures only part of the retained storefront-authenticity need and remains underspecified relative to the closest verified target. Selecting Ask is an action hit; full request-alignment credit is not automatic. Credit is assigned by the best-matching retained target, so the model need not enumerate every item that later appeared.}
        \label{fig:case-finriskask-partial}
    \end{figure}
    
    \begin{figure}[!htbp]
        \centering
        \includegraphics[width=\linewidth]
        {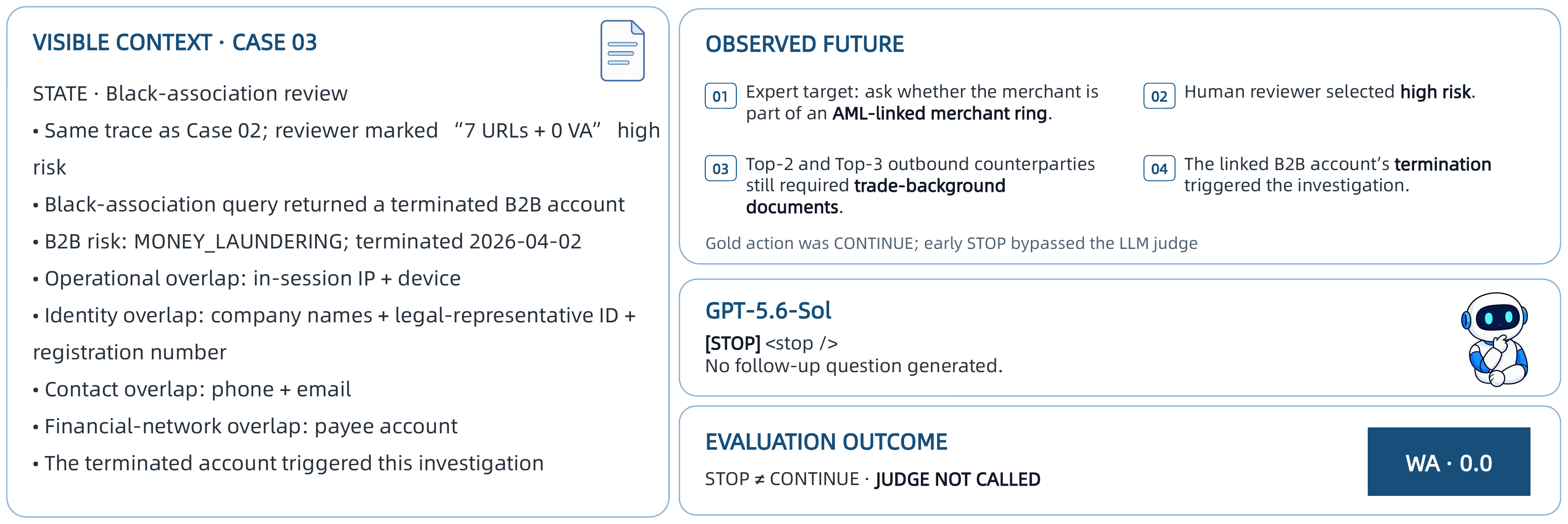}
        \caption{\textbf{Missed information acquisition ($z_t=0$).}
        The recorded state is Ask, but the model selects Proceed and issues no request. Because the Ask branch is not entered, no candidate request is compared with the retained target set and the state contributes zero to ERA. The state is an Ask-recall error even if proceeding from the visible record appears locally plausible.}
        \label{fig:case-finriskask-missed-acquisition}
    \end{figure}

    An Ask prediction with an invalid or unrelated request receives zero request-alignment credit despite entering the correct branch. Conversely, a Proceed prediction on a recorded Ask state fails to enter request evaluation because no evidence request is produced. These cases illustrate why action agreement, Ask-state coverage, and request targeting are reported separately in FinRisk-Ask.
    
\section{Limitations and Responsible Use}
\label{app:limitations}

    \paragraph{Scope and interpretation.}
    FinRiskAtlas focuses on Chinese-language financial risk-control and compliance review and evaluates model configurations under explicitly defined operation contracts. The reported scores should therefore be interpreted as measurements of performance on the released review operations and evidence states, rather than as a general certification of financial-review capability across all institutions, jurisdictions, or workflows. The benchmark is designed to study how evaluation units influence model selection in professional review settings.
    
    \paragraph{Evidence-state evaluation boundary.}
    FinRisk-Ask evaluates evidence-state control through offline replay of completed professional trajectories. The benchmark measures whether a model follows the recorded transition and whether its request aligns with trajectory-supported evidence needs under the released evaluation protocol. Because targets are derived from evidence realized in completed trajectories, FinRisk-Ask evaluates alignment with verified evidence needs rather than executing requests in a live review environment or estimating their downstream operational impact.
    
    \paragraph{Professional reference behavior.}
    The Ask-or-Proceed labels represent observed reviewer transitions from one enterprise risk-control workflow. They provide a consistent reference boundary for evaluation while preserving the fact that professional workflows may differ across organizations, policies, and operating conditions. Accordingly, FinRisk-Ask measures agreement with the released workflow behavior and request-target alignment, rather than defining a universal review policy.
    
    \paragraph{Data provenance and annotation.}
    FinRiskAtlas combines benchmark instances constructed from professional materials, regulatory resources, financial and legal records, and de-identified review trajectories. Static benchmark families and FinRisk-Ask states were defined and reviewed by domain experts with financial risk and legal backgrounds. Expert review was applied to taxonomy design, family contracts, instance quality control, action mapping, and evidence-target verification. Raw enterprise trajectories are not released; only de-identified reconstructed states and evaluation artifacts are provided.
    
    \paragraph{Responsible use.}
    FinRiskAtlas is intended as an evaluation resource for studying and comparing model configurations in professional financial-review scenarios. It is not designed to replace qualified reviewers or support autonomous financial or legal decisions. Deployment in real workflows should additionally consider institution-specific policies, regulatory requirements, human oversight, and operational constraints beyond the benchmark contracts.

\end{document}